\documentclass[12pt]{article}
\usepackage{lscape}
\usepackage{geometry}
\usepackage[onehalfspacing]{setspace}
\usepackage{amssymb}
\usepackage{amsfonts}
\usepackage{graphicx}
\usepackage{amsmath}
\usepackage{epstopdf}
\usepackage{natbib}
\usepackage{bibunits}
\usepackage{morefloats}
\usepackage{float}
\usepackage{comment}
\usepackage{pdflscape}
\usepackage{adjustbox}
\usepackage{subcaption}
\usepackage{longtable}
\usepackage{apalike}
\usepackage{xfrac}
\usepackage[dvipsnames,table]{xcolor}
\usepackage{etoolbox,siunitx}
\robustify\bfseries

\usepackage{qtree}
\usepackage{multicol}
\usepackage{booktabs}
\usepackage{threeparttable}
\usepackage{tabularx}
\usepackage{dcolumn}
\newcolumntype{d}[1]{D..{#1}} % for alignment of numbers on decimal marker
 
\usepackage{siunitx}
\usepackage{mathpazo} % math & rm
\usepackage[scaled]{helvet} % ss
\usepackage{courier} % tt
\normalfont
\usepackage[T1]{fontenc}
\usepackage{listings}
\usepackage{epigraph}
\def\sym#1{\ifmmode^{#1}\else\(^{#1}\)\fi}

\usepackage{xcolor} % Include the package
\definecolor{DarkerPineGreen}{RGB}{0, 90, 80} % Define a custom darker shade
 
\definecolor{darkblue}{RGB}{10, 10, 100}

\usepackage{hyperref}
\usepackage{rotating}

\hypersetup{
  colorlinks,
  citecolor=DarkerPineGreen,
  linkcolor=red,
  urlcolor=blue}
\usepackage{cleveref}

\usepackage{algpseudocode}

\DeclareMathOperator*{\argmin}{arg\,min}
 
\usepackage{etoolbox}
 
\makeatletter
\patchcmd{\epigraph}{\@epitext{#1}}{\itshape\@epitext{#1}}{}{}
\makeatother
 
\renewenvironment{abstract}
 {\small
  \begin{center}
  \bfseries \abstractname\vspace{-.5em}\vspace{0pt}
  \end{center}
  \list{}{
    \setlength{\leftmargin}{1.6cm}    \setlength{\rightmargin}{\leftmargin}  }  \item\relax}
 {\endlist}
\usepackage{graphicx}
\usepackage{wrapfig}
 
\allowdisplaybreaks
 
\usepackage{enumitem}

\begin{document}
\sloppy
\title{\vspace*{-0cm} \Huge To Bag is to Prune} 
\author{Philippe Goulet Coulombe\thanks{%
Département des Sciences Économiques,  \href{mailto:p.gouletcoulombe@gmail.com}{\texttt{goulet\_coulombe.philippe@uqam.ca}}.  For helpful discussions and/or  comments,  I would like to thank Karun Adusumilli, Edvard Bakhitov, Christophe Barrette,  Frank Diebold, Maximilian G{\"o}bel, Samuel Gyetvay, Stephen Milborrow, Frank Schorfheide, Dalibor Stevanovic, David Wigglesworth and Boyuan Zhang. For excellent research assistance, I thank  Hugo Couture, Akshay Malhotra and Tony Liu. }}
\date{\vspace{-0.4cm}
Université du Québec à Montréal\\[2ex]%
\small
\small
First Draft: July 10, 2020 \\
This Draft: \today \\ 
%\href{https://drive.google.com/file/d/1W_-t5EOn3ulLz6SVOVs1AHHGrc7SAb4h/view?usp=sharing}{Latest Draft Here} \\ %\today
\vspace{0.3cm}
\large
%{\color{PineGreen} Preliminary and incomplete: \\
% please do not circulate without permission.} \\
  }%  Most Recent Version \href{https://www.dropbox.com/s/zdoepjbu2w9h4vu/PGC_RF.pdf?dl=0}{Here}
\maketitle
%Finally, most of this paper was written during a stay in Saint-Anne-des-Monts, QC. I'm thankful to all the great and inspiring people I met there.
 
\begin{abstract}

\noindent It is notoriously difficult to build a bad Random Forest (RF). Concurrently, RF blatantly overfits in-sample without apparent consequences out-of-sample. Arguments like the bias-variance trade-off or double descent cannot rationalize this paradox. I propose a new explanation: bootstrap aggregation and model perturbation as implemented by RF automatically prune a latent "true" tree.  More generally,  I document that randomized ensembles of greedily optimized learners implicitly perform optimal early stopping \textit{out-of-sample}.  So, letting RF overfit the training data is a dominant tuning strategy against nature's undisclosed choice of noise level.   Additionally,  novel ensembles of Boosting and MARS are also eligible.  I empirically demonstrate the property, with simulated and real data, by reporting that these new completely overfitting ensembles perform similarly to their tuned counterparts --- or better.  
 
%This makes RF particularly reliable in macroeconomic forecasting applications where the use of cross-validation is not without perils.
 
%This makes RF particularly reliable in forecasting applications where the use of cross-validation is not without perils.
 
%So,  letting RF completely overfit the training sample is a dominant tuning strategy against nature's choice of noise level.  This makes RF particularly reliable in macroeconomic forecasting applications where the use of cross-validation is not without perils. Additionally,  novel ensembles of Boosting and MARS are also eligible for optimal implicit early stopping.  I empirically demonstrate the property, with simulated and real data,  by reporting that these new completely overfitting ensembles perform similarly to their tuned counterparts --- or better.  

%They work particularly well for macroeconomic forecasting where the use of cross-validation is not without perils.

\end{abstract}
 
\thispagestyle{empty}
%\noindent \textit{JEL Classification: C53, C55, E37, C32} 
 
%\noindent \textit{Keywords: Random Forest, Trees, Regime Switching, Structural Breaks, Time-varying Parameters}
 
%EndExpansion
 
%\noindent \textit{JEL codes: C32, C55, E27}
 
%\noindent \textit{Keywords: Vector Autoregressions, LASSO, Ridge, Time-varying Parameters, Factor models}
 
\clearpage
 
%\thispagestyle{empty}
%\tableofcontents
%\thispagestyle{empty}
 
\clearpage 
\setcounter{page}{1}
 
\section{Introduction}
 
Random Forest (RF) is a very stubborn benchmark in Machine Learning (ML) applications to economic and financial forecasting.  It can successfully predict asset prices \citep{kellyml}, house prices \citep{mullainathanspiess2017}, and macroeconomic aggregates \citep{medeiros2019,chen2019off,GCLSS2018,MDTM}. It can infer treatment effect heterogeneity \citep{athey2019grf}, and estimate generalized time-varying parameters \citep{MRF}. The list goes on. But what makes it so distinctively reliable? To answer that question, and eventually understand the reasons behind RF's growing list of successful applications, it is better to start with an apparent paradox.
 
\begin{figure}[h!] %{0.5\textwidth}
\vspace*{-0.2cm}
  \begin{center}
\includegraphics[trim={1.75cm 0cm 0.5cm 0cm},clip,scale=.35]{{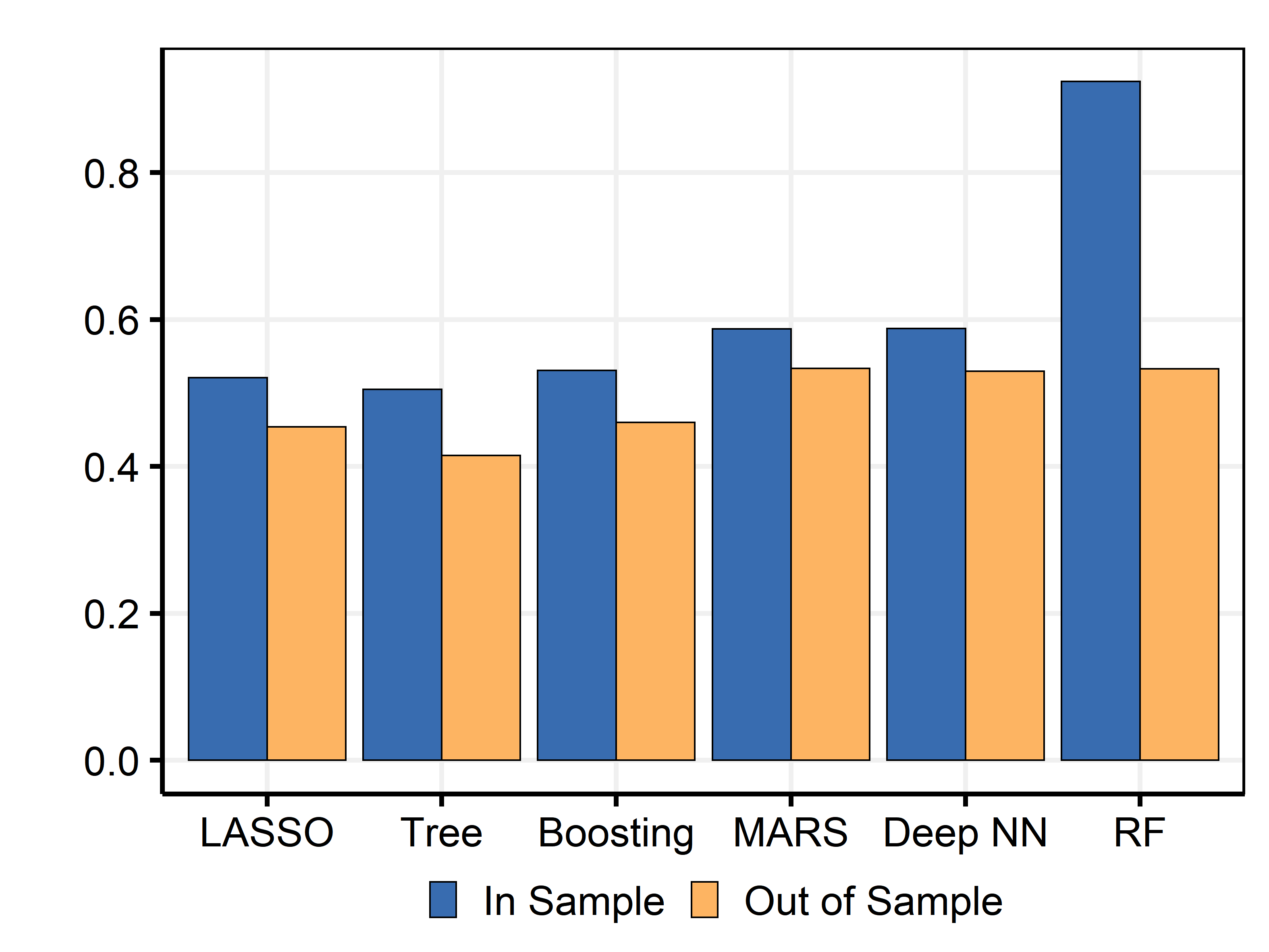}}  
\vspace{-0.2cm}
\caption{\footnotesize \href{http://archive.ics.uci.edu/ml/datasets/Abalone}{\textit{Abalone}} data set: comparing $R^2_{\text{train}}$ and  $R^2_{\text{test}}$. First four models hyperparameters are tuned by 5-fold CV. RF uses default parameters. NN details are in Appendix \ref{sec:nndetails}.}
\label{rinrout}
    \end{center}
  \vspace*{-0.25cm}  
\end{figure}
 
Common statistical wisdom suggests that a non-overfitting supervised learning algorithm should have approximately the same mean squared error in the training sample as in the test sample. LASSO, Splines, Boosting, most\footnote{This is not always true of deep learning applications because of a phenomenon called "Double Descent". This will be discussed and contrasted with RF later in the paper.} Neural Networks (NN), and Multivariate Adaptive Regression Splines (MARS)  abide by that principle. But not Random Forest. RF typically has an exceptionally high in-sample $R^2$ with a much lower, yet competitive, out-of-sample one.\footnote{Throughout the paper, the coefficient of determination of model $j$ is $R^2_{j}=1-\sfrac{MSE_j}{MSE_{\bar{y}}}$ where $\bar{y}$ is the mean of $y$.  Then,  $R^2_{\text{train}}$ means the classic in-sample $R^2$ while $R^2_{\text{test}}$ means both model $j$ and $\bar{y}$ predictions are trained,  then  projected on new data.  For $\bar{y}$, this means the average of the training set is the prediction.  Unlike $R^2_{\text{train}}$,  $R^2_{\text{test}}$ can be negative when the more complex model ends up being worse than the simple mean prediction.  The use of $R^2$ is preferred because we are in a regression environment and it provides a standardized way of looking at performance that, unlike ratios of MSEs, keeps relevant information about the signal-to-noise ratio in a data set -- a key aspect for this paper.} This means not only do the individual trees overfit the training set, but that the ensemble does, too. In contrast, the algorithms mentioned above usually perform poorly in such conditions. When optimally tuned, they are expected to deliver neighboring  $R^2_{\text{test}}$ and $R^2_{\text{train}}$. Figure \ref{rinrout} testifies to all those observations. This paper is about understanding why RF is excused from obeying the $R^2_{\text{test}} \approx R^2_{\text{train}}$ rule — and showing how to leverage this property for other algorithms.
 
Providing a theoretical reason to believe RF will not overfit, \cite{breiman2001} shows that the generalization error is bounded. That bound goes down as the individual learners’ strength increases and goes up as correlation between them increases.  Despite recent theoretical advances, like proving consistency \citep{scornet2015consistency}, it is still unclear why RF works so well on so many data sets.  It is acknowledged that much of that resilience is attributable to RF providing a flexible non-linear function approximator that does not overfit.  Most importantly, unlike many models of the nonparametric family, the latter characteristic seems guaranteed even without resorting to careful hyperparameters tuning.

%Yet, it is still not clear what mechanism is behind this phenomenon.
 
If RF -- made of fully grown completely overfitting trees -- does not overfit out-of-sample, where does regularization come from? Clearly, increasing $\lambda$ (the tuning parameter guiding the weight on the $l_2$ norm penalizing parameters) brings regularization in a ridge regression by shrinking coefficients toward zero, lowering the individual importance of each predictor. When it comes to RF, what contortions on the intrinsic model does its regularization entail? An appealing answer is that bagging smooths hard-thresholding rules \citep{buhlmann2002analyzing}, like increasing the smoothness parameter of smoothing splines. If that were the whole story, RF, as does smoothing splines, would yield comparable $R^2_{\text{test}}$ and $R^2_{\text{train}}$. Model averaging arguments would also have a similar implication.\footnote{This renders incomplete (at best) arguments linking RF regularization to that of penalized regression (originally discussed in \cite{ESL}, and more recently \cite{mentch2019randomization}) using results developed for \textit{globally optimized} linear models \citep{elliott2013complete,lejeune2020implicit}.} As clearly displayed in Figure \ref{rinrout}, it is not the case --- so something else must be at work.\footnote{\cite{mullainathanspiess2017}’s Table 1 -- reporting results from off-the-shelf ML algorithms applied to house price prediction --  is another convenient example where all aspects of the phenomenon are visible.} The newly proposed hypothesis is: to bag (and perturb) is to prune.   More precisely,  bagging many trees that completely overfit $y_{\text{train}}$ will be shown to be empirically equivalent to bagging "oracle" pruned trees -- i.e., trees whose depth is known to be optimal after looking at the test set.  
 
%\textbf{\color{red} NEW}
%It is well known that bagging reduces variance
 
%That is,  the  But not a CART-based tree.  Rather, RF implicitly prunes a latent true underlying tree.  
%\textbf{central contribution} is to 
 
More generally,  this paper empirically documents that randomized greedy optimization (of parameters in statistical models) performs optimal early stopping.\footnote{Greedy optimization in this context refers to models which are constructed sequentially/adaptively by doing the myopic/heuristic next best step (as in,  e.g.,  Forward Stagewise Regression) rather estimating all parameters of a completely specified model by minimizing the sum of squared residuals from that fixed structure through a gradient (like ordinary least squares (OLS)). } This is interesting since greedy optimization is often introduced in statistical learning books as an inevitable (but suboptimal) practical approach in the face of computational adversity \citep{ESL}. It turns out the necessary evil has unsuspected benefits. A greedy algorithm treats what has already happened as given and what comes next as if it will never happen. While this depiction usually means "trouble", it is the key to this paper’s argument. By recursively fitting a model and \textit{not} re-evaluating what came before as the algorithm progresses, the work of early stages will be immune to subsequent overfitting steps, provided the latter averages out efficiently. Mechanically, when running CART, the structure at the top (like the already chosen cutting variables and values) cannot be weakened by the bottom’s doings -- the bottom’s existence is not even considered when estimating the top. Moreover, when faced with only noise left to fit in a terminal node, it is shown that a \textit{Perfectly Random Forest}’s test set prediction is the sample mean, which is unbiased and -- most importantly -- has minimal variance.  Said differently,  in this specific context, it gives an identical out-of-sample prediction as that of the optimally pruned model.  
 
Fortunately, not only trees are eligible for the enviable property, but also other greedily fitted additive models like Boosting and MARS \citep{MARS}. Based on this observation, I develop novel variants coined \textit{Booging} and \textit{MARSquake} which -- like RF -- are ensembles (of bagged and perturbed base learners) that completely overfit the training sample and yet perform nicely on the test set (which is typically not observed for plain Boosting and MARS).  Those are later shown to be promising alternatives to Boosting and MARS (both with a tuned stopping point) on real and simulated data sets. An \href{https://github.com/philgoucou/bagofprunes}{\texttt{R} package} implements both.  Nonetheless,  those are not introduced to provide yet another marginal twist on Boosting and MARS,  but to further demonstrate that no early stopping (the number of trees in Boosting and pruning in MARS) is needed when ensembling any (properly) randomized greedy learners. Thus,  the conditions for RF's robustness live beyond it -- even when there are no trees involved.
 
%\textbf{\color{red} NEW} 
 
%\footnote{In fact, Booging can be seen as a special case of \cite{pfahringer2000winning} where one deliberately let the number of trees grows so large that each base learner has perfect fit.} 
 
\vskip 0.15cm
{\noindent \sc \textbf{Summary of Contributions.}}   They are fourfold.  First, the paper brings to light RF's large and unique -- yet still undocumented -- wedge between $R^2_{\text{test}}$ and $R^2_{\text{train}}$, making standard explanations of RF's success  incompatible with it or incomplete.  Second,  it argues through intuitive explanations and experimental evidence (on both simulated and real data) that this is due to ensembles of randomized \textit{greedy} optimization performs optimal early stopping -- a previously unknown phenomenon.  Third,  in the case of RF,   it is showcased that this translates to RF pruning some form of latent tree object, and thus RF's $R^2_{\text{test}}$ being monotonically increasing with trees depth.  Fourth, simple new algorithms are introduced to showcase that aforementioned observations about randomized greedy algorithms are not specific to RF.   %that is, they apply even when no tree whatsoever is involved.
 
%\textbf{\color{red} NEW}
 
%[1] Pfahringer, Bernhard. "Winning the KDD99 classification cup: Bagged boosting." ACM SIGKDD Explorations Newsletter 1.2 (2000): 65-66.
 
\vskip 0.15cm
{\noindent \sc \textbf{Organization.}}  This paper is organized as follows. In section \ref{sec:rf_miracle}, I present the main insights and discuss their implications for RF and other greedy algorithms.   Additionally,  the section contains a comparison of this paper's explanation with recent "interpolating regime" and "double descent" ideas proposed to explain the success of deep learning \citep{belkin2019reconciling}.  In section \ref{sec:simuls}, I demonstrate by means of simulations the implicit optimal early stopping property of RF, Booging and MARSquake. Section \ref{sec:empirics} applies the paper’s ideas to classic regression data sets. Section \ref{sec:con} concludes.

\section{Randomized Greedy Optimization \& Optimal Early Stopping}\label{sec:rf_miracle}
 
This section first introduces the Random Forest algorithm in detail and then proceeds to formulate this paper's main arguments.
 
\subsection{Random Forest Algorithm}\label{sec:primer}

Random Forest (RF) is a diversified ensemble of regression trees.  I first introduce regression trees and present their estimation via a greedy algorithm,  then discuss the ensembling procedure.   
 
\vskip 0.15cm
{\noindent \sc \textbf{A Tree.}} Consider the cross-sectional scenario where $\pi_i$ represents the inflation for country $i$, and $r_i$ denotes the nominal rate of interest for country $i$.  In addition, let $g_i$ be a measure of the output gap.  A simple decision tree for the cross-sectional distribution of inflation rates at a given point in time could look like 
 
\vspace{1em}
\Tree[.{Full Sample} 
[.{$g_i \geq 0 $} 
[.{$r_{i} \geq 4 \%$} {$\phantom{--} \pi_i= 2 +\epsilon_i \phantom{--}$} ]
[.{$r_{i}  < 4 \% $} {$\phantom{--} \pi_i = 5 +\epsilon_i \phantom{--}$} ]
 ]
[.{$g_i  < 0 $} {$\phantom{--} \pi_i = 1 +\epsilon_i \,  .     \phantom{--}$} ]
 ]
 \vspace{1em}
 
 \noindent This can be expressed in the form of a regression equation: $y_i = {T}(X_i) + \epsilon_i$, where ${T}$ represents the tree and $X_i$ a vector of features for instance $i$.    Once its structure is fixed,   $T$ can be written as a linear regression where features created from $X$ are interactive dummies.  For instance,  the above tree could be formulated as 
\begin{align}\label{regeq}
y_i = 1  \times  I(g_i < 0) + 5 \times I(g_i  \geq 0 )I(r_{i}  < 4) + 2 \times I(g_i  \geq 0 )I(r_{i} \geq 4) + \epsilon_i.
\end{align}
In practice,  both the values and the dummies to be used are unknown.  Whatever the representation,  the central algorithmic question remains the same: how does one explore the vast model space spanned by any interactions of many dummies created from many variables at various cutoff points.  
 
\vskip 0.15cm
{\noindent \sc \textbf{Greedy Estimation.}} The usual strategy -- introduced as Classification and Regression Trees (CART) in \cite{breiman1984classification} --  is to deploy a \textit{greedy} algorithm that \textit{recursively} partitions the data according to
\begin{equation}\label{treesplit}
\begin{aligned}
\min\limits_{k\in \mathcal{K}, \smallskip c \in {\rm I\!R}}\Bigg[ \min\limits_{\mu_{1}} \sum \limits_{\{ i \in L| X_{i,k} \leq c  \}} \left(y_{i}-\mu_{1}\right)^{2} 
 + \min\limits_{\mu_{2}} \sum \limits_{\{ i \in L | X_{i,k} > c  \}} \left(y_{i}-\mu_{2}\right)^{2}\Bigg].
\end{aligned}
\end{equation}
Here, $\min\limits_{k\in \mathcal{K}, c \in {\rm I\!R}}$ denotes the minimization over all possible splits, 
where $k$ indexes a variable in $\mathcal{K}$, representing available features,  and $c$ is a real number representing the split point.  $L$ is a leaf,  representing the sub-sample of data utilized by \eqref{treesplit} to estimate the next split.  The first $L$ in the recursion is the whole training sample,  then the algorithms proceeds recursively by using the subsamples created by the previous partition as the subsequent $L$'s in the next iteration.  This mechanically creates partitions of ever-decreasing size until we reach a stopping point (which needs to set according to some rule) and obtain a set of terminal nodes.  For instance,  the simplistic inflation tree example above features 3 terminal nodes.  
 
The overall goal of a single step is to find the optimal pair $( k^*,  \,  c^*)$ and the predicted values ($\mu_{1},  \,  \mu_{2}$)  that minimize the total within-leaf sum of squared errors.  In this context,  $\mu_{1}$ and $\mu_{2}$  are always the within-leaf sample average.  While local optimization is obtained \textit{de facto},  there is no guarantee that successively running \eqref{treesplit},  which chooses the pair $( k^*,  \,  c^*)$ taking previous splits as given and ignoring future ones, will deliver the globally optimal tree.  This is typical of  greedy algorithms,  a broad set of methods that employs the heuristic of selecting the \textit{locally} optimal solution at \textit{each step} of the problem-solving process.  This sequential model-building scheme  is computationally fast but makes for an inevitably fragile $\hat{T}$.   Thereby,  a single tree with many splits,  while accommodating of complex data generative processes, tends to have prohibitively high variance.  Tree variance can be mitigated by pruning, which involves grouping terminal nodes together by deleting insignificant splits that occurred at  a later stage of the tree-building process.  But this has many  limitations,  the pruned tree is still a locally optimized structure and the extent of pruning is a tuning parameter on which predictive performance heavily relies on.   
 
\vskip 0.15cm
{\noindent \sc \textbf{Diversifying the Portfolio.}} A strategy that has been found to be much more successful in machine learning practice is that of creating a diversified portfolio of trees,  seeing each tree $\hat{T}$ as a base learner,  and take the average of their predictions.  The process from turning a single tree into forest can be summarized as three main ingredients.
 
\begin{enumerate}
\item[\texttt{\large D }:  ]  First,  we need to let each of the constituent trees run \textit{deep},  that is, having a plethora of terminal nodes obtained from trees of great depth coming from a long sequence of splits.   Even though this implies near-complete overfitting if the base learner were not meant to be ensembled,  we run \eqref{treesplit} until all terminal nodes contain very few observations (usually less than 5).  This ingredient makes sure bias is minimized by allowing for constituents trees of great complexity.  As we will later,  the absence of a trade-off for the depth parameter -- and thus the usual wisdom of always using deep trees -- is a direct consequence of the implicit pruning phenomenon described in this paper.
 
\item[\texttt{\large B }:  ]  Second,  we conduct \textit{Bagging},  which stands for Bootstrap Aggregation  \citep{breiman1996bagging}.  We create $B$ nonparametric bootstrap samples of the data,  picking $[y_i \enskip X_i]$ pairs with replacement.   Each tree in the forest is estimated using a bootstrapped sample $b \in 1,\dots, B$.  When the base learner exhibits high non-linearity in observation and/or instability, the benefits of Bagging are significant \citep{breiman1996bagging}. 
 
\item[\texttt{\large P }:  ] Third,  the building process is subject to \textit{perturbation} via stochastic constraints.  Since a Random Forest is essentially an average of numerous trees, the variance reduction from averaging is more substantial when the component trees are uncorrelated.  At each $L$ where \eqref{treesplit} is computed,  we only consider a subset of all predictors ($\mathcal{K}^{-} \subset \mathcal{K}$) for the split.  This compels the algorithm to explore a variety of optimization paths, forcing it to avoid consistently choosing identical splits across trees  (especially at the upper levels of the $\hat{T}_b$).  This strategy ensures that the greedy algorithm, which is inherently local in its optimization, does not follow the same route for every $b$.  Consequently, the trees become more diversified and the computation time is reduced by evaluating fewer pairs at each application of \eqref{treesplit}.  The fraction of predictors randomly selected,  typically referred to as \texttt{mtry}  is a tuning parameter with a default value of $\sfrac{1}{3}$ in a regression context \citep{breiman2001}.
\end{enumerate}
 
\noindent RF prediction for new instance $j$ is the simple average of all the $B$ tree predictions : $\hat{y}_j^{\text{RF}} = \sfrac{1}{B}\sum_{b=1}^B \hat{T}_{b}(X_j)$.  In terms of classical bias-variance view of RF,  \texttt{\large D} brings bias down, while \texttt{B} \& \texttt{P} mitigate variance.  This section provides a new understanding of how that mitigation takes shape. 
 
\vskip 0.15cm
{\sc \noindent \textbf{Preliminary Observations on the Behavior of RF}.} As a consequence of \texttt{D},  it is common to see that RF will have $R^2_{\text{train}}$  magnitudes higher than $R^2_{\text{test}}$, a symptom which would suggest overfitting for many standard algorithms. That is,  the traditionally defined in-sample fitted values ($\hat{y}_i^{\text{RF}} = \sfrac{1}{B}\sum_{b=1}^B \hat{T}_{b}(X_i)$) and corresponding residuals have nothing to do with what one gets when applying the estimated model to new data -- unless the “true” $R^2$ is really high.  %\footnote{Everything in this paper assumes a $B$ large enough so that averages are stable \citep{ESL,scornet2017tuning} since $B$ is not a tuning parameter in RF (unlike Boosting).}
 
While this $R^2_{\text{train}}$ curiosity is usually of limited interest \textit{per se}, it creates some intriguing headaches from a more traditional statistical perspective. For instance, any attempt to interpret the intrinsic RF \textit{model} relies on measurements obtained on pseudo hold-out samples (called \textit{out-of-bag}). In contrast, one would not refrain from exploring the structure of Lasso’s fitted values or that of a single tree. Indeed, most algorithms, when properly tuned, will produce comparable $R^2_{\text{test}}$ and $R^2_{\text{train}}$. This implies that using the in-sample conditional mean $\hat{y}_i$ for any subsequent analysis is perfectly fine. In that way, they behave similarly to any classical nonparametric estimators where a bandwidth parameter must be chosen to balance estimation flexibility and the threat of overfitting. Once it is chosen according to CV or some information criteria, in-sample values provide reliable estimates of the \textit{true} conditional mean and error term. 
 
I argue that RF’s notably different behavior can be explained by the combination of two elements: greedy optimization and randomization of the recursive model fitting sequence through \texttt{B} \& \texttt{P}.  By construction, the instability of trees  makes the latter an easy task: simply bootstrapping the original data can generate substantially different predictors \citep{breiman1996bagging}. The former, greedy optimization, is usually seen as the suboptimal yet inevitable approach when solving for a global solution is computationally intractable.  In this section, I argue that greedy optimization, when combined with randomization of the model building pass, has an additional benefit. When combined in a  properly randomized ensemble, no harm will come in letting each greedily optimized base learner ($\hat{T}_{b}$ in RF) completely overfit the training sample. In the case of RF,  this translates to the heuristic recommendation of \texttt{D} -- i.e.,  considering fully grown trees where each terminal node contains either a single observation or very few.  Subsequently,  those observations are leveraged to develop new algorithms inheriting RF's desirable properties.

\subsection{What Happens in the Overfitting Zone Stays in the Overfitting Zone}\label{sec:ofz}
 
Unlike trees,  a linear regression model can be estimated greedily \textit{and} globally (that is called OLS), so they are a natural example to contrast the merits of greedy vs. global methods.  Moreover, as explicitly laid out in Section \ref{sec:expend},  a tree can be rewritten as some sort of additive/multiplicative linear model.
 
In a global estimation procedure, overfitting will weaken the whole prediction function.  More concretely, estimating many useless coefficients in a linear regression will inflate the generalization error by increasing the variance of \textit{both} the few useful coefficients and the useless ones. Bagging such a model will still be largely suboptimal: the ensemble relies on an average of coefficients which are largely inferior to those that would be obtained from regression excluding the useless regressors.  An identical issue arises in Complete Subset Regression (CSR, \citealt{elliott2013complete}),  which conduct model averaging of many regression models with different subsets of regressors.  In CSR, we still need to tune the number of included regressors,  since including too many will substantially deteriorate performance (for instance, see \cite{KLS2019} 's empirical results).  Hence, we are still in the standard case where $R^2_{\text{test}}<R^2_{\text{train}}$ reveals that the model's performance is inferior to that of an optimally pruned counterpart.
 
%Call the recursively constructed function $\hat{f}$.
A greedily optimized model works differently.  At each optimization step, everything that came before is treated as given and what comes next as if it will never happen.  That is, as the algorithm progresses past a certain step $s$, the function estimated before $s$ is treated as given.  And everything before $s$ was estimated assuming anything past $s$ to be non-existent.  This is a direct consequence of the algorithms decisions being only locally optimal.  Eventually, the greedy algorithm will reach $s^*$ where the only thing left to fit is the unshrinkable “true” error $\epsilon_i = \hat{\epsilon}_{i,s^*} = y_i - \hat{f}_{s^*}(x_i)$.  The key is that entering deep in the overfitting zone will not alter $\hat{f}_{s-1}$ since it is not re-evaluated.  As a result, early non-overfitting steps can be immune to the weakening effect of subsequent ones, as long as the latter efficiently averages out to 0 in the hold-out sample.  An immediate implication of this separability property is that there is no need to stop the model building sequence at the unknown $s^*$ to obtain  predictions immune to estimation variance inflation -- provided what happens past $s^*$ averages out efficiently.  As we will see for RF in section \ref{sec:prf},  it does.
 
These abstract principles can be readily applied to think about fitting trees where a step $s$ is splitting the subsample obtained from step $s-1$.  A tree does not distinguish whether the current sample to split is the original data set of the result of an already busy sequence of splits.  Moreover, like any splits along the tree path, those optimized before venturing past $s^*$ cannot be subsequently revoked. This implies that the predictive structure attached to them cannot be altered nor weaken by subsequent decisions the greedy algorithm makes. 
 
\begin{figure} %{0.5\textwidth}
\vspace*{-0.7cm}
  \begin{center}
\includegraphics[trim={1.25cm 0.005cm 0cm 0cm},clip,width=0.5\textwidth]{{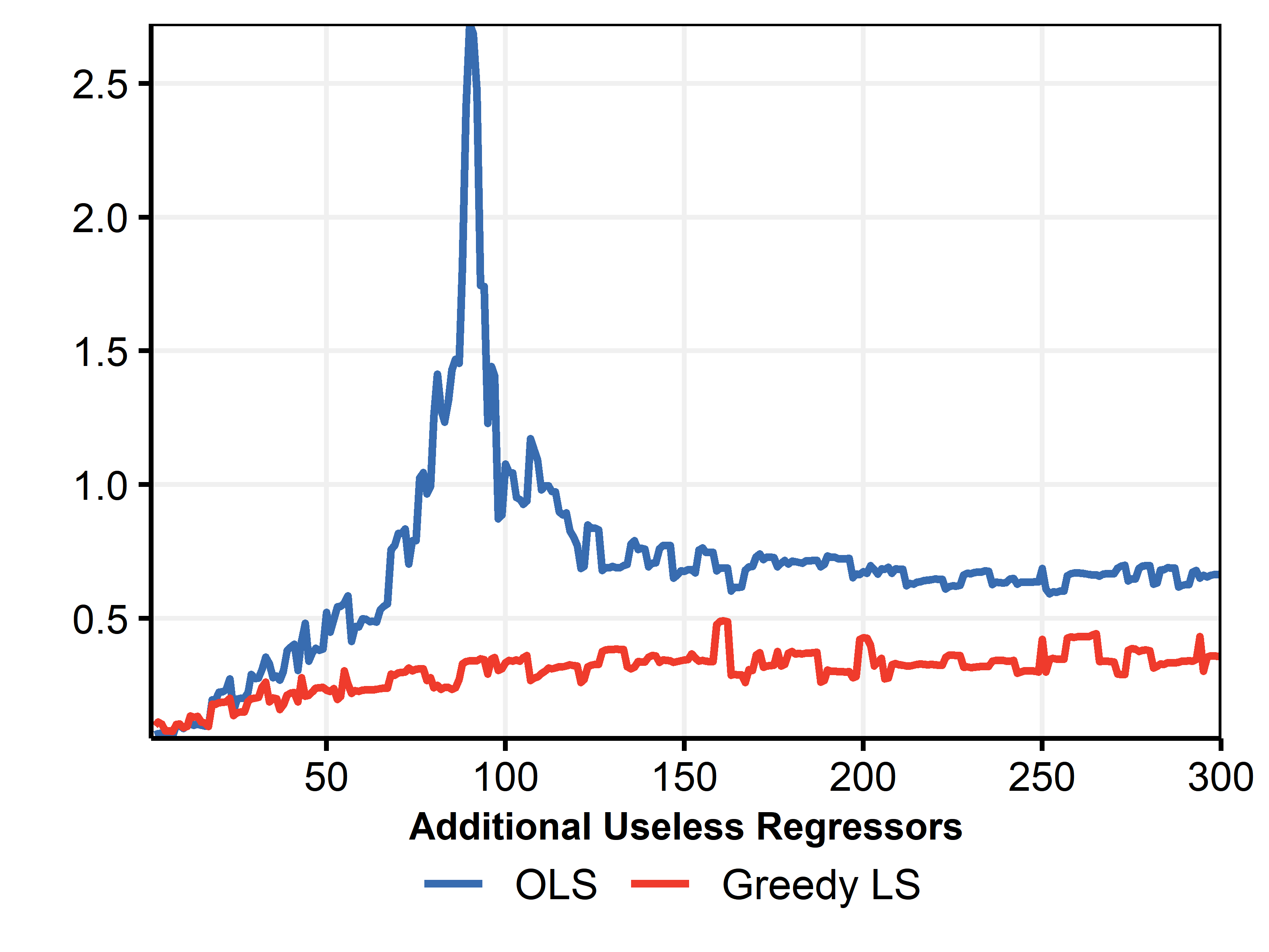}}  %NN_vs_RF_compact}} %
\vspace{-0.25cm}
\caption{\footnotesize Model averaging/bagging different base learners with increasingly many useless features. Units are $\ln{(\sfrac{\text{MSE}_{\text{model}}}{\text{MSE}_\text{Oracle}})}$. Oracle has 10 regressors, SNR=2, and $N=100$.  Each prediction is the average over 50 models and 20 bagging replicas. For OLS, bagging is bypassed since it provides the same expectation as using the full sample once. The 50 models are constructed as follows: for each model, I generate $x$ new useless regressors (features made of noise that are not entering the DGP) and add them to the relevant ones, then run estimation.  "Greedy OLS"  is  \texttt{glmboost} in \texttt{R}, setting the learning rate at 1.}
\label{modavg}
    \end{center}
  \vspace*{-0.4cm}  
\end{figure}
   
Alternatively, we can think of fitting a linear regression with orthogonal features. A step $s$ is adding a regressor by fitting it to the residual of the previous step. In this linear boosting case, we can hope that important predictors go in very soon in the process and are followed by many useless predictors until those are exhausted. Unlike the coefficients from the kitchen-sink OLS (throwing in the regression every predictor we have), the early fitted coefficients in the model building sequence of the stagewise algorithm were estimated as part of a model that only included a handful of predictors. Those are precluded from the eventual weakening effect that comes with the inversion of a near-singular $X’X$.  Adding a ridge penalty will alleviate the singularity problems, but will also (potentially heavily) shrink the real coefficients of interest, compromising their predictive power.

Figure \ref{modavg} supports those observations: for the same linear model, the effects of model averaging and bagging can differ substantially.  Clearly,  Greedy LS (linear boosting setting the learning rate at 1)  responds much better to ensembling than OLS in an environment incorporating noise and useless regressors.  OLS's performance past the interpolation threshold ($R^2_{\text{train}}=1$, which occurs at 90) is still order of magnitudes worse than Greedy LS (the graph is in log scale).  There are limited benefits from tuning (moving along the $x$-axis) in Greedy LS, while those are huge for OLS.\footnote{Considering random subsets of regressors -- or equivalently adding $l_1$ or $l_2$ regularization -- to help OLS performance is tempting yet beside the point: we want to see how the two methods behave in an increasingly hostile environment (useless regressors) when the level of hostility is unknown beforehand (and hence the optimal regularization).  We see that while cross-validation can greatly help OLS,   its potential benefits for Greedy LS are limited. }

%\footnote{Moreover, OLS has a slight built-in advantage because it always includes the relevant regressors plus useless ones whereas Greedy LS has to select them itself.  It is understood that OLS's performance would deteriorate even further without this crutch. }
 
%Adding a ridge penalty will alleviate singularity problems, but will also shrink coefficients of interest, compromising their predictive power. 
 
\subsection{Bagging and Perturbing as an Approximation to Population Sampling}\label{sec:prf}
 
As a byproduct of the key observation discussed in the previous section,  we can conduct mathematical analysis by looking at different $s$'s separately.  One such $s$ of interest to establish "to bag is to prune" in certain contexts is $s^*$, which corresponds to the \textit{true} terminal node in the case of a tree.  Past $s^*$,  we enter the overfitting zone -- that is,  fitting noise -- and whatever the algorithm does next will hurt its ability to generalize.   At that point, the data generating process (DGP) is simply 
\begin{align}\label{dgp}
y_{i, s=s^*}= \mu +\epsilon_i
\end{align}
where $\epsilon_i$ is the true error (i.e,  $\epsilon_i=y_{i} - E(y_{i}|X_i)$).  To alleviate notation,  $y_{i, s=s^*}$ will be referred to as $y_{i}$ from now on.  Clearly, the best possible prediction in this environment is the mean of all observations contained in the node.  Under this DGP,  any tree model more complex than the sample average can be pruned without any increase in bias.  I argue that perfect randomization (that is,  \texttt{B} \& \texttt{P} generate uncorrelated trees when confronted to white noise) will also procure this optimal prediction out-of-sample, even if the ensemble itself is completely overfitting in-sample. This \textit{Perfectly} Random Forest is, of course, merely a theoretical device and how close RF gets to this hypothetical version is an empirical question.  It is showcased in section \ref{sec:simuls}) that bagging (\texttt{B}) and perturbing (\texttt{P}) trees can get very close to what one would get from population sampling, an empirical device designed to obtain perfect randomization in a simulated data environment.  Essentially,  each tree is grown on non-overlapping samples from a population, which is the ideal experiment that the bootstrapping of any statistic is meant to approximate.  Of course, the performance of any model will improve when averaging it over many close-to independent samples. The more subtle point being made here is that a good approximation to population sampling (via \texttt{B} \& \texttt{P}) can generate a model whose structure will be close to the optimally pruned one, and that, without attempting any form of early stopping whatsoever. In other words, \eqref{dgp} more generally represents the truth from the hypothetical point $s^*$ where a recursive fitting algorithm should optimally stop. Proper inner randomization assures that a prediction close to $\bar{y}$ is returned.  
 
This can be formalized to provide intuition in a more sanitized environment.   The derivations uses fully grown trees which contain one observation in each node (this implies that each base learners’ $R^2_{\text{train}}$ is one).  Precisely,  under the DGP in  \eqref{dgp},  the out-of-sample prediction from a Perfectly Random Forest made of fully grown trees coincides with the optimally pruned prediction.  The out-of-sample prediction of a RF for observation $j$ is
$$\hat{y}_j^{\text{RF}}=\frac{1}{B} \sum_{b=1}^B \hat{y}_{j,b} $$
where $\hat{y}_{j,b}$ is the prediction of the $b$ tree for a new observation $j \notin \texttt{train}$.  The perfect randomization assumption (which is legitimate in the context of \eqref{dgp}) implies that each out-of-sample tree prediction is a randomly chosen $y_i$ for each $b$. The prediction is thus
$$\hat{y}_j^{\text{RF}}= \frac{1}{B} \sum_{b=1}^B y_{i(b)}.$$
Define $r = \sfrac{B}{N}$ where $N$ is the number of training observations and $r$ as “replicas”. Since the $y_{i(b)}$’s amount to random draws of $y_{1:N}$, for a large enough $B$, we know with certainty that the vector to be averaged ($y_{i(1:B)}$) will contain $r$ times the same observation $y_i$. Hence, the prediction equivalently is
$$\hat{y}_j^{\text{RF}}= \frac{1}{B} \sum_{i=1}^N \sum_{r’=1}^r y_{i,r’}=\frac{1}{B} \sum_{i=1}^N \sum_{r’=1}^r y_{i}=\frac{r}{B} \sum_{i=1}^N y_{i}=\frac{1}{N} \sum_{i=1}^N y_{i}$$
since $r=\sfrac{B}{N}$. 
 
%   say simple mean is the optimal
%
 
%\begin{lemma}
 
%\end{lemma}
 
%\begin{proof}
 
%\end{proof}

%Importantly, observation $j$ is not included when fitting the trees, so we are looking at the prediction for a new data point using a function trained on observations $i \neq j$.   
 
%Just say B+P can be appromitated by pure randomization where faced with epsilon
 
In words, when a Perfectly Random Forest -- which can be obtained "empirically' from Population Sampling as described in Section \ref{sec:simuls} -- is starting to fit pure noise, its out-of-sample prediction averages out to the simple mean, which is \textit{optimal} under \eqref{dgp} and a squared loss function. Intuitively, at $s^*$, the test set behavior of the prediction function (from fully grown trees) is identical to that of doing (random) subsampling with subsamples containing one observation. Averaging the results of the latter (over a large $B$) is just a complicated way to compute a mean. Hence, the out-of-sample prediction as provided by the perfectly random forest is one where implicit/automatic pruning was performed.\footnote{This provides a justification for \cite{duroux2016impact}'s finding that pruning the base learners while shutting down $\texttt{B}$ can deliver a performance similar to that of RF (provided a wise choice of tuning parameters).} It is equivalent to that of an algorithm which knows the “true” $s^*$. A direct implication is that we need not to worry about finding $s^*$ through cross-validation, since the optimally stopped prediction is what is being reported out-of-sample. Of course, this relies on a satisfying randomization level to be empirically attainable. Section \ref{sec:simuls} asks “How close to population sampling are we when fitting \texttt{B} \& \texttt{P} trees?” and the answer is “very close”.  
 
%\footnote{This also directly implies that each base learners’ in-sample predictions ($\hat{\mu}_{i,b}$) correlation with $y_i$ is one and that of the ensemble $\sfrac{1}{B} \sum_{b=1}^B \hat{\mu}_{i,b}$ is bounded below by the subsampling rate ($Corr(\sfrac{1}{B} \sum_{b=1}^B \hat{\mu}_{i,b},y_i)=\sfrac{1}{B} \sum_{b=1}^B Corr(  \hat{\mu}_{i,b},y_i)>\sfrac{1}{B}\sum_{b=1}^B E ( 1 \times I(i \in \texttt{train}) + 0 \times I(i \in \texttt{train})) = \texttt{subsampling rate}$),  which will inevitably be much higher than the true correlation of 0.  } 
 
%When faced with only noise left to fit, the prediction of a \texttt{B} \& \texttt{P} ensemble of completely overfitting trees achieving perfect randomization is the (optimal) sample average. This result is obtained from taking the recursive view and assuming perfect randomization. 

%Below, I complete the “pruning” argument by showing that RF is plausibly being pruned is rather a \textit{latent} tree. 
 
%Population sampling itself does not generate $R^2_{\text{test}} < R^2_{\text{train}}$\footnote{As the subsampling rate mechanically decrease -- a luxury obtained from a growing sample size and fixed model complexity -- the $R^2_{\text{train}}$ itself will look like $R^2_{\text{test}}$ (thus, reflecting the real signal-to-noise ratio) since the contribution of observation $i$ to its out-of-bag prediction shrinks with subsampling rate.}, only its approximation by \texttt{B} \& \texttt{P} does. 
 
The above also helps in understanding $R^2_{\text{test}} < R^2_{\text{train}}$ in RF.  The gap’s existence is a direct implication of implicit pruning via \texttt{B} \& \texttt{P} being only active \textit{out-of-sample}.  A central role in this is that of $\texttt{mtry}$ -- the number of randomly selected features to be considered for a split.  Overfitting arise from an overabundance of (unregularized) parameters vs. observations. The attached predictors are either directly available in the data or created via some form of basis expansions which trees is one possibility out of many.  In such high-dimensional situations, it is clear that the model itself -- the predictive structure -- is barely identified: many different tree structure can rationalize a training sample with $R^2_{\text{train}}=1$. Yet, these structures' predictions substantially differ when feeding in new data. This property of overfitting models (combined with the recursive fitting procedure) is the channel through which $\texttt{mtry}$ strongly regularize the hold-out sample prediction. However, the resulting heterogeneity cannot deflate $R^2_{\text{train}}$ since \textit{different} overfitting base learners, when trained on the same data, provide the \textit{same} fitted values ($y_{\text{train}}$ itself).  Ergo, $R^2_{\text{test}} < R^2_{\text{train}}$.

\subsubsection{An Analytical Example}\label{sec:ana}
 
Suppose that we have reached the point of $y_i = \mu + \epsilon_i$ and assume that  $\epsilon_i \sim iid N(0,\sigma^2)$.   Furthermore,  we have at our disposal $B$ regressors $X_{i,b} \sim iid N(0,\cdot)\,  \,  \forall (i,  b)$.   That is,  features are useless because independent of $y_i $ and are independent across themselves (for simplicity).  Finally,  suppose,  for tractability,  that we have 4 observations,  3 for training, and the fourth one for test.   The choice of 3 observations is the minimal number of observations to have splits which are not all delivering the same MSE in-sample (splitting two observations always deliver a MSE of 0 after the split).  With 3 observations,  we can have a node with 2 observations and one with a single one.  Thus, from the perspective of \eqref{treesplit}, there is an actual optimization choice to be made.   From this ensues three possible partitions of the training data:  \[
\mathcal{P}_1 = \left( \left\{ y_1, y_2 \right\}, \left\{ y_3 \right\} \right), \quad 
\mathcal{P}_2 = \left( \left\{ y_1, y_3 \right\}, \left\{ y_2 \right\} \right), \quad 
\text{and} \quad 
\mathcal{P}_3 = \left( \left\{ y_2, y_3 \right\}, \left\{ y_1 \right\} \right).
\]
A partition can be delivered by more than a single $X_{b}$.  In fact,  with a large $B$,  we expect numerous   $X_{b}$ to deliver, say $\mathcal{P}_1$ ,  by $I(X_{i,b}> 0)$.\footnote{For simplicity,  I assume in this example that all considered splits occur at $c=0$.} This point is crucial,  as it is directly related to the diversification potential of RF:  while numerous $X_{b}$'s  deliver observationally equivalent partitions in-sample,  their allocation on the test sample will differ depending on the draws of $X_{4,  b}$ (i.e., for the out-of-sample observation).

%($X_{i,A}> 0$,  $X_{i,B }> 0$,  $X_{i,C}> 0$),  that is,  $\mathcal{K} = \{1,2,3\}$ with $c=0$, that delivers three possible partitions of the training data : 
\vskip 0.15cm
{\noindent \sc \textbf{Composition of the Forest.}} A first question we ask is the prevalence of each of these partitions being the outcome of  \eqref{treesplit} in this context.  The probability of partition $\mathcal{P}_1$ being favored to $\mathcal{P}_2$ and $\mathcal{P}_3$ depends on \textit{in-sample} MSEs:
\begin{align}
\operatorname{P} (\mathcal{P}_1 \succ \mathcal{P}_2 \,  \, \cap \,  \,  \mathcal{P}_1 \succ \mathcal{P}_3 ) =  \operatorname{P}( \text{MSE}_{\mathcal{P}_1} < \min\{\text{MSE}_{\mathcal{P}_2}, \text{MSE}_{\mathcal{P}_3}\})
 \end{align}
 where 
 \begin{align*}
 \text{MSE}_{\mathcal{P}_1} &= \left(\frac{y_1+y_2}{2}-y_1\right)^2 + \left(\frac{y_1+y_2}{2}-y_2 \right)^2 +(y_3 - y_3)^2 = \frac{(y_1-y_2)^2}{4} \\
  \text{MSE}_{\mathcal{P}_2} &= \left(\frac{y_1+y_3}{2}-y_1\right)^2 + \left(\frac{y_1+y_3}{2}-y_3 \right)^2 +(y_2 - y_2)^2 = \frac{(y_1-y_3)^2}{4}  \\
   \text{MSE}_{\mathcal{P}_3} &= \left(\frac{y_2+y_3}{2}-y_2\right)^2 + \left(\frac{y_2+y_3}{2}-y_3 \right)^2 +(y_1 - y_1)^2 = \frac{(y_2-y_3)^2}{4} \, \, .
 \end{align*}
which is simply  $\text{MSE}_{\mathcal{P}_j} = \sum_i^3 (\hat{y}_{i,j} -y_i)^2$ for $j \in \{ 1,2, 3\}$ and $\hat{y}_{i,j}$ are fitted values implied by $\mathcal{P}_j$.  Thus,  the expected share of trees delivering $\mathcal{P}_1$ in this example can be obtained through
 \begin{align}\label{proba}
  \operatorname{P}( \text{MSE}_{\mathcal{P}_1} < \min\{\text{MSE}_{\mathcal{P}_2}, \text{MSE}_{\mathcal{P}_3}\}) = \operatorname{P} ((y_1 - y_2)^2 < \min((y_2 - y_3)^2, (y_3 - y_1)^2)) \, \, .
 \end{align}
Calculating \eqref{proba} analytically is difficult  because the squared differences of these variables do not follow a standard $\chi_2$ distribution. It is trivial to simulate,  however.  The result is,  as expected,  $\sfrac{1}{3}$ for a large enough simulation sample and this probability is invariant to either $\mu$ or $\sigma^2$.  Thus,  each partition gets an \textit{equal weight} and thus,  if there are $B$ trees in total for a large enough $B$,  we expect there would be $\sfrac{B}{3}$ replications of each partition, or said differently,  there will be $\sfrac{B}{3}$ trees delivering $\mathcal{P}_j$ for $j \in \{ 1,2, 3\}$.   For what follows,  it assumed that each of those $\sfrac{B}{3}$ trees giving $\mathcal{P}_j$ are based on splits utilizing different $X_b$ for each $b$.  For instance,  while $X_b$ and $X_{b'}$ may both deliver $\mathcal{P}_1$ in-sample and therefore the same MSE,  their decision for whether the out-of-sample observation $y_4$ gets predicted with $\hat{y}_{4,b}=\frac{y_1+y_2}{2}$ or $\hat{y}_{4,b}=y_3$ depends on the realizations of $X_{4,b}$ and $X_{4,b'}$.  Those will likely differ in the context of fitting noise with mutually uncorrelated and useless regressors. 
 
The equal share of $\sfrac{1}{3}$ for each partition obviously  hinges on the null DGP at $s^*$.   Deviating from it gives different probabilities,  and results dependent on $\sigma^2$.   For instance,  when the means ($\mu$) are adjusted to 3 for \( y_1 \) and \( y_2 \), and to 1 for \( y_3 \), the estimated probability increases to about 60.48\%.  Altering the means to -3 for \( y_1 \) and \( y_2 \), and to 1 for \( y_3 \), the estimated probability further increases to approximately 89.85\%.   Thus,  when there is actual signal,  evidence piles up and trees concentrate, when there is not,  competing partitions are equally weighted. 
 
\vskip 0.15cm
{\noindent \sc \textbf{Comparison of Predictions With the Mean.}}  Now that we know the weight of each of the partitions,  we can compare \textit{out-of-sample} predictions of RF  for a hypothetical $y_4$ versus the (optimal) mean under the usual null DGP.  First,  from $\operatorname{E}\left[  \hat{y}^{\text{RF}}_4 \right]= \operatorname{E}\left[  \hat{y}^{\text{mean}}_4 \right] = \mu $ we have that
\begin{align*}
\operatorname{E}\left[ \left( \hat{y}^{\text{RF}}_4 - \hat{y}^{\text{mean}}_4 \right)^2 \right] &=  \operatorname{Var} \left[ \hat{y}^{\text{RF}}_4 - \hat{y}^{\text{mean}}_4  \right]  
\end{align*}
Then, we can input the definitions of the two predictions schemes as a function of the decisions rules based on  $X_{4,b}$,  the weights of $\sfrac{1}{3}$ derived above,  and the corresponding predictions computed from training data partitions. 
\begin{align*}
\operatorname{Var} \left[ \hat{y}^{\text{RF}}_4 - \hat{y}^{\text{mean}}_4  \right]  &=\operatorname{Var}\left[  \frac{1}{B}  \sum_{b = 1}^B T_b(X_{4,b})  - \frac{y_1+y_2+y_3}{3}   \right] \\ 
&= \operatorname{Var}\left[  \frac{1}{B} \Bigg( \sum_{b = 1}^{\sfrac{B}{3}} T_b(X_{4,b}) + \sum_{b=\sfrac{B}{3} +1}^{\sfrac{2B}{3}} T_b(X_{4,b})+ \sum_{b = \sfrac{2B}{3} +1}^{{B}} T_b(X_{4,b}) \Bigg) - \frac{y_1+y_2+y_3}{3}    \right] \\ 
&= \operatorname{Var}\left[  \frac{1}{B} \Bigg(  \sum_{b = 1}^{\sfrac{B}{3}} \left( I(X_{4,b}>0)\left(\frac{y_1+y_2}{2} \right)  + I(X_{4,b}\leq 0) y_3 \right) \right. \\
&\quad \quad \quad+ \left. \sum_{b=\sfrac{B}{3} +1}^{\sfrac{2B}{3}} \left( I(X_{4,b}>0)\left(\frac{y_1+y_3}{2} \right)  + I(X_{4,b}\leq 0) y_2 \right) \right. \\
&\quad \quad \quad + \left. \sum_{b = \sfrac{2B}{3} +1}^{{B}} \left( I(X_{4,b}>0)\left(\frac{y_2+y_3}{2} \right)  + I(X_{4,b}\leq 0) y_1 \right) \Bigg) - \frac{y_1+y_2+y_3}{3}    \right]
\end{align*}
For compactness,  let $D_b \equiv I(X_{4,b}>0)$ and thus,   $ I(X_{4,b}\leq 0) = 1 - D_b$.  Then,   given the aforementioned assumptions on $X_{b}$,   we have that  $D_b \sim iid \,  \operatorname{Bernoulli}(\frac{1}{2})$.  Moving forward by grouping $D_b$ terms together, we get
\begin{align*}
&= \operatorname{Var} \left[  \frac{1}{B}  \left( \sum_{b = 1}^{\sfrac{B}{3}} D_b\left(\frac{y_1+y_2}{2} - y_3\right) + \sum_{b=\sfrac{B}{3} +1}^{\sfrac{2B}{3}} D_b\left(\frac{y_1+y_3}{2} - y_2\right) + \sum_{b = \sfrac{2B}{3} +1}^{B} D_b\left(\frac{y_2+y_3}{2} - y_1\right) \right)  \right.  \\  &\quad \quad \enskip \smallskip \,  \,  \left.  + \frac{1}{B}\left( \frac{B}{3}y_3 + \frac{B}{3}y_2 + \frac{B}{3}y_1\right) - \frac{y_1+y_2+y_3}{3} \right] \\
&= \frac{1}{B^2} \operatorname{Var}\bigg[ \underbrace{ \left(\frac{y_1+y_2}{2} - y_3\right) \sum_{b = 1}^{\frac{B}{3}} D_b + \left(\frac{y_1+y_3}{2} - y_2\right) \sum_{b=\frac{B}{3} +1}^{\frac{2B}{3}} D_b + \left(\frac{y_2+y_3}{2} - y_1\right) \sum_{b = \frac{2B}{3} +1}^{B} D_b}_{\equiv \, Z} \bigg]
\end{align*}
\noindent At this juncture,  we want to use the law of total variance by conditioning on $y_1$, $y_2$, and $y_3$ written compactly as $\boldsymbol{y}$.  Thus, the above expression can be decomposed such that
\begin{align*}
     \frac{1}{B^2} \operatorname{Var}\left[ Z \right] =  \frac{1}{B^2} \operatorname{Var}_{\boldsymbol{y}} \left[ \operatorname{E}\left[ Z \middle\vert \boldsymbol{y} \right] \right] +  \frac{1}{B^2} \operatorname{E}_{\boldsymbol{y}} \left[ \operatorname{Var}\left[ Z \middle\vert \boldsymbol{y} \right] \right] 
     \,  .
\end{align*}
Noting that $\operatorname{E}(D_b)= \frac{1}{2}$ under our assumptions,   the first term becomes
\begin{align*}
    &= \frac{1}{B^2} \operatorname{Var}_{\boldsymbol{y}} \left[ \left(\frac{y_1+y_2}{2} - y_3\right) \frac{B}{3}  \frac{1}{2}  + \left(\frac{y_1+y_3}{2} - y_2\right) \frac{B}{3}  \frac{1}{2} + \left(\frac{y_2+y_3}{2} - y_1\right)\frac{B}{3}  \frac{1}{2} \right] \\
&= \frac{1}{B^2} \times \frac{B^2}{36} \operatorname{Var}_{\boldsymbol{y}} \left[  \left(\frac{y_1+y_2}{2} - y_3\right) + \left(\frac{y_1+y_3}{2} - y_2\right) + \left(\frac{y_2+y_3}{2} - y_1\right) \right] \\
&= \frac{1}{36} \operatorname{Var}_{\boldsymbol{y}} \left[ \frac{y_1}{2} + \frac{y_1}{2} - y_1 + \frac{y_2}{2} + \frac{y_2}{2} - y_2 + \frac{y_3}{2} + \frac{y_3}{2} - y_3 \right] = \frac{1}{36} \times 0 \, .
\end{align*}
This is intuitive given that one can easily show that,  beyond $\operatorname{E}\left[  \hat{y}^{\text{RF}}_4  \right]= \operatorname{E}\left[  \hat{y}^{\text{mean}}_4 \right] = \mu$,   we have  $\operatorname{E}\left[  \hat{y}^{\text{RF}}_4 \vert \boldsymbol{y} \right]= \operatorname{E}\left[  \hat{y}^{\text{mean}}_4 \vert \boldsymbol{y} \right]$ from the earlier calculation about the composition of the forest --- and we are here calculating their difference.  Moving to the second term,  with  $\operatorname{Var}(D_b)= \frac{1}{2} \times \left(1 - \frac{1}{2} \right) = \frac{1}{4}$ and under the $iid$ assumptions,  it collapses to 
\begin{align*}
   &= \frac{1}{B^2} \operatorname{E}_{\boldsymbol{y}} \left[  \left(\frac{y_1+y_2}{2} - y_3\right)^2 \frac{B}{3} \frac{1}{4}  + \left(\frac{y_1+y_3}{2} - y_2\right)^2 \frac{B}{3} \frac{1}{4}  + \left(\frac{y_2+y_3}{2} - y_1\right)^2 \frac{B}{3}  \frac{1}{4}  \right] \\
      &=   \frac{1}{12 B} \operatorname{E}_{\boldsymbol{y}} \left[  \left(\frac{y_1+y_2}{2} - y_3\right)^2 + \left(\frac{y_1+y_3}{2} - y_2\right)^2 + \left(\frac{y_2+y_3}{2} - y_1\right)^2  \right] \\
      &= \frac{3}{12 B} \operatorname{E}_{\boldsymbol{y}} \left[  \left(\frac{y_1+y_2}{2} - y_3\right)^2   \right] = \frac{3 }{8} \times  \frac{ \sigma^2 }{ B} \,  .
\end{align*}
The middle step leverages linearity of expectation then identical distributions for each of the squared terms.  Thus,  putting it all together,  we get that $\lim_{B\rightarrow \infty } \operatorname{E}\left[ \left( \hat{y}^{\text{RF}}_4(B) - \hat{y}^{\text{mean}}_4 \right)^2 \right]  = \lim_{B\rightarrow \infty } \frac{3 }{8} \times  \frac{ \sigma^2 }{ B} = 0$.  In the scenario of a  perfectly random forest, where a substantial number \( B \) of independent trees exists, the prediction made by the Random Forest (RF) aligns with that of the mean.  This is analogous to an oracle forecast, which is aware of having attained the optimal stopping point \( s^* \) and consequently delivers the most accurate prediction possible in this context.  However, when applying this concept to practical situations, it is essential to interpret \( B \) used in this derivation cautiously.  Here, \( B \) refers to the number of independent trees rather than the quantity of trees selected by the user for aggregation. Under the assumptions made, \( B \) is more aptly understood as an effective number of independent trees.  In cases where the random forest has a limited number of regressors, resulting in little opportunities for diversification and thereby correlated trees, the effective number of independent trees \( B \) could substantially fall short of their official count (or simulated draws).  Therefore,   diversification is crucial.  We will see from the simulations in Section \ref{sec:simuls} that the performance of RF often closely approaches that of a perfectly randomized scheme, suggesting the above derivations,  although for a stylized RF,  are very indicative of its true  behavior.   
 
Finally,  note that this simple analytical example obeys the $R^2_{\text{train}} \geq R^2_{\text{test}}$ regularity.  $R^2_{\text{test}}$ is 0 whereas $R^2_{\text{train}}>0$.  Yet,   in the null DGP,  0 is the best attainable $R^2$ out-of-sample.  If RF were overfitting,  $R^2_{\text{test}}$ would be negative.

\subsection{Leveraging the Insight for Other Models}\label{sec:expend}
 
%\color{darkblue}{

Not only trees require pruning or some form of early stopping.  Many additive schemes must be optimally stopped at $s^*$ to obtain the best test set performance. It has been discussed that mixing \texttt{B} \& \texttt{P} with a greedy recursive algorithm can lead the algorithm to perform implicit optimal early stopping.  It is natural to wonder if certain well-known greedy model building algorithms could also benefit for this property.   
 
\vskip 0.15cm
{\noindent \sc \textbf{Overview of Boosting and MARS.}} A generic algorithm of the Boosting family,  minimizing the sum of squared errors loss  can be summarized as follows.
\vspace{-0.4em}
\begin{description}[itemsep=0pt, parsep=0pt,leftmargin=2cm]
    \item[\phantom{------}\textbf{Given:}] Training set \( \{(X_1, y_1), \ldots, (X_N, y_N)\} \)
    \item[\phantom{------}\textbf{Initialize:}] \( F_0(X) = \bar{y} \) or 0
    \item[\phantom{------}\textbf{For} \( s = 1 \) \textbf{to} \( S \):]
    \begin{enumerate}[itemsep=0pt, parsep=0pt, leftmargin=*]
    \item[\phantom{o}]
        \item Compute residuals: \( r_{is} = y_i -  F_{s-1}(X) \)
        \item Fit a regression tree to the residuals:  \( g_{s}(X) = \arg \min_{g} \sum_{i=1}^N (r_{is} - g(X))^2 \)
        \item Update the model: \( F_s(X) = F_{s-1}(X) + \nu g_s(X) \)
    \end{enumerate}
    \item[\phantom{------}\textbf{Output:}] \( F_S(X) = \sum_{s' =0}^S \nu g_{s'}(X) \) if \( S \) is reached or some criterion is met
\end{description}
Let $g$ be a tree $T$ of limited depth and we have the most basic Boosted Trees algorithm for regression.    Boosting is a method that {{recursively}} combines forecasts from many over-simplistic trees, often referred to as weak learners, into a strong learner.  It begins with fitting a shallow tree (e.g.,  with $\texttt{depth}$ typically ranging from 1 to 5), which is a weak predictor with a large bias in the training sample.  The method then \textit{iteratively} adds trees that fit the prediction residuals of the existing ensemble.  The procedure is {greedy} because each $g$ choice at substep 2 is optimized myopically,  taking the preceding sequence as given and not considering upcoming optimization steps.  Each new tree's forecast is shrunken by a factor $\nu \in(0,1)$ to prevent rapid overfitting. The process continues until $S$ trees form the ensemble.  The final output is an additive model of shallow trees with three key tuning parameters: $(\texttt{depth} ; \nu ; S)$. Performance improvements can be achieved through various enhancements,  such as subsampling $r_{is}$ (and $X_{i}$) via stochastic Gradient Boosting \citep{friedman2002}.  While there is no pruning going on in the classical sense in standard Boosting algorithm,  the cross-validation of the key tuning parameter $S$ plays a similar role.  Preferably,  the chosen $S$ should be close to the theoretical optimal stopping point $s^*$,  which here,  also depends on $\nu$.
 
Multivariate Adaptive Regression Splines (MARS) can also be cast within the above algorithm by setting $\nu=1$ and $g_s(X) = \beta_s  h_s (X) $ where $\beta_s$ are coefficients and $h_s$ are basis functions.  $h_s$'s are composed of constants or hinge functions,  which are piecewise linear and defined as either $\text{max}(0, x - t)$ or $\text{max}(0, t - x)$,  where $x$ represents an independent variable and $t$ is a knot.  Note that while each hinge function uses a single column of $X$ at a time,  MARS also considers $h_s$'s which are interactions of many such hinge functions using different variables or knot placement,  up to a prespecified maximal degree of interaction.  The estimation algorithm consists of two main steps, the forward pass and the backward pass.  First,  starting with an intercept,  MARS \textit{iteratively} adds terms to the model that most reduce the residual sum of squares. This process continues until adding further terms does not significantly improve the fit, or a predefined maximum number of terms is reached. Then,  the backward pass can proceed:  the model is  pruned by stepwise removal of the least effective terms, based on criteria such as Generalized Cross-Validation.  Thus, like CART,  MARS is a built through a greedy algorithm and overfitting is controlled by peeling off the layers that were added last in the forward pass.  The level of pruning is decided according to a pseudo out-of-sample metric \citep{MARS,earthpackage}.   
 
%\footnote{For those unfamiliar with this machinery, see \cite{friedman2002} for Boosting and \cite{MARS} or \cite{earthpackage} for MARS.} 
 
\vskip 0.15cm
{\noindent \sc \textbf{New Self-Pruning Boosting and MARS Ensembles.}} Following the observation that both Boosted Trees and MARS are built through greedy algorithms and are thus eligible for implicit optimal pruning,   I  report the performance of  \texttt{B} \& \texttt{P} versions of (overfitted) MARS and Boosted Trees in sections \ref{sec:simuls} and \ref{sec:empirics} -- that is,  on synthetic and real data,  respectively.   The objective is to compare how they fare versus optimally stopped versions where $S$ is tuned using usual strategies.  \texttt{B} is implemented for both MARS and Boosting via running many times the algorithms on $B$ subsamples of the data without replacement.  \texttt{P} is implemented in the same fashion as RF for MARS, by stochastically restricting which variable can be used to enter the hinge function at each substep 2 in the algorithm above.  For Boosting,  \texttt{P} is most easily implemented by activating Stochastic Gradient Boosting,  which creates its own kind of perturbation by stochastically restricting the training observations to be used to build $T$ in substep 2, therefore perturbing the variables choice sequence beyond what  \texttt{B} does for the complete model building pass.

Additionally,  the considered \texttt{B} \& \texttt{P} versions of MARS and Boosting will feature the so-called data augmentation (\texttt{DA}) option activated.  It consists in enlarging the feature matrix to additionally incorporate $\tilde{X}=X+\mathcal{E}$ where $\mathcal{E}$ is a matrix of Gaussian noise.  For categorical variables, $\tilde{X}$ is obtained by duplicating $X$ and shuffling a fraction of its rows.  Overall, \texttt{DA} can improve perturbation's potential when regressors are scarce and bring the algorithm closer to ideal of perfect randomization -- a key ingredient for bagging to result in out-of-sample pruning.  The \textit{overfitted} Boosting and MARS \texttt{B} \& \texttt{P} + \texttt{DA} versions will be referred to by the sobriquets \textit{Booging} and \textit{MARSquake}. An \href{https://github.com/philgoucou/bagofprunes}{\texttt{R} package} implements both.   All the relevant technicalities and execution details to effectively implement \textit{Booging} and \textit{MARSquake} from the usual  \textit{Boosting} and \textit{MARS} packages are relegated to Appendix \ref{sec:details}.

\vskip 0.15cm
{\noindent \sc \textbf{Limitations of Randomized Boosting and MARS Ensembles.}} A recurring theme in section \ref{sec:simuls} is that while Randomized Boosting and MARS Ensembles benefit from implicit pruning as well,  the wedge between an optimally randomized version and the empirical one is sometimes wider for those than Random Forest -- for which the gap is virtually non-existent.  Also, we see signs of minor degradation in performance past $s^*$ in noisier environments for the two alternatives ensembles.   I discuss here why Boosting and MARS  can plausibly benefit from implicit pruning, but perhaps not as much as trees. The success of randomized greedy algorithms is bounded by the base learner's ability to generate a sufficiently diversified ensemble of predictors.  If not, there will be benefits from stopping base learners earlier.  %The derivations in section \ref{sec:ana} made that clear through how variance 
 
%It is to have the greedy base learner generates enough randomization so the ensemble consists of diversified predictors. 

The reason why trees generate the most randomization among greedy methods is the \textit{irreversibility} of the model building pass.  The clearest example is surely that of trees: once the dataset has been segmented according to a predictor and a cutting values,  there is no turning back.  Plain Boosting is used as the counterexample here, but the principle clearly applies to MARS and similar greedily optimized  additive models. Consider building a small symmetric tree of depth 2, the prediction function is
\begin{align*}
\hat{y}_i &=  I(x_{i}>0)\left[\alpha_1 I(z_{i}>0)+\alpha_2 I(z_{i}\leq 0)\right] + I(x_{i}\leq 0)\left[(\gamma_1 I(w_{i}>0)+\gamma_2 I(w_{i}\leq 0)\right].
\end{align*}
Finally, define $d_{x,i}^{+}=I(x_{i}>0)$ as a regressor and the rest accordingly. We get
\begin{align*}
\hat{y}_i = \theta_1 d_{x,i}^{+} d_{z,i}^{+}+\theta_2  d_{x,i}^{+} d_{z,i}^{-} +\theta_3 d_{x,i}^{-}d_{w,i}^{+}+\theta_4 d_{x,i}^{-} d_{w,i}^{-}.
\end{align*}
This representation shows trees are very singular "additive" models. It is clear that $d_{x,i}$ better be a good choice, because it is not going away: any term in the model building pass will be multiplied by it. By construction, no term added later in the expansion has the power to entirely undo the damage of a potentially harmful first split. In other words, splitting the sample  is an \textit{irreversible} action. This is what guarantees that steps occurring past $s^*$ will not alter what was constructed before it.
 
Now, let us look at a toy boosting model where the base learners are single-split trees (stumps) and the learning rate is $\nu$. An important difference with the above is that step $s$ leading to 
\begin{align}\label{boosting_correction}
\hat{y}_i &= \nu \left[\beta_1 d_{x,i}^{+}+\beta_2 d_{x,i}^{-}\right] + \dots -\nu \left[\beta_1 d_{x,i}^{+}+\beta_2 d_{x,i}^{-}\right]
\end{align}
is absolutely possible. In words, by additivity, it is possible to correct any step that eventually turned out to be suboptimal in the search for a close-to global optimum.  With the randomization induced  through \texttt{P} + \texttt{DA},  this is unlikely to happen exactly in those terms.  Nevertheless, a small $\nu$ and a large number of steps/trees in the additive model will mechanically increase the algorithm's potential for "reversibility".  Indeed, \cite{rosset2004boosting} detail an equivalence between a procedure similar to the above and LASSO. If $\nu \rightarrow 0$, $ \text{\# of steps}\rightarrow \infty$ and regressors are uncorrelated, they obtain the LASSO solution -- a \textit{global} solution. Thus, there is an imminent tension between how close to a global optimization \eqref{boosting_correction} can get and its capacity to generate inner randomization sufficiently to be dispensed from tuning the stopping point. %This is intuitive. There is only one truth to be learned, and learning it slowly will safely get you there. Alternatively, by learning fast and imprecisely, you get it right on average over many tryouts. 
 
An interesting question is whether the properties detailed here apply to LASSO, which would free the world from ever tuning $\lambda$ again. Indeed, when implemented via Least Angle Regression \citep{efron2004least}, the algorithm very much looks like a forward stagewise regression. In the spirit of the above, one would hope to let a randomized version of the regularization path roll until $\lambda=0$, average those solutions and obtain the same $R^2_{\text{test}}$ as if $\lambda$ had been carefully tuned. Unfortunately, LASSO violates two of the requirements listed before.  First, parameters are re-evaluated along the regularization path. For $\lambda$'s that lay in the overfitting territory, the estimated coefficients will be weakened since they are re-estimated in an overcrowded model. Second, letting the model overfit (when $K<N$) implies setting $\lambda=0$ which returns the OLS solution for any iteration, making the desired level of randomization likely unattainable.  This last point could be alleviated, when in the high-dimensional $K>N$ case, the LASSO solution can include at most $N$ predictors.  In that scenario, the included set of variables would depend on the order within the regularization path (rather than its termination) which would increase randomization.  Nevertheless, we cannot expect LASSO to benefit from automatic tuning because linear regression coefficients are re-evaluated along the estimation path.
 
\subsection{Why RF is Not Equivalent to Pruning a Single Tree}\label{sec:setar}

It is well known that,  as a result of randomization, RF performs orders of magnitude better than a single pruned tree \citep{breiman1996bagging}.   This is also observed in the simulations from section \ref{sec:simuls}: \texttt{B} \& \texttt{P} CART does  better than the ex-post optimally pruned base learner, \textit{even when the DGP is not smooth}.  In contrast, \texttt{B} \& \texttt{P} MARS and Boosting will provide similar performance to that of their respective base learner stopped at $s^*$.  Thus, RF is pruning something, it must be something else than CART.  I complete the argument of previous sections by proposing the hypothesis that "pruning via inner randomization" is applied on the true \textit{latent} tree $\mathcal{T}$ in 
\begin{align}\label{eq:truetree}
y_i = \mathcal{T}(X_i) + \epsilon_i
\end{align}
which itself can only be constructed from randomization.  In short,  this necessity is due to the imperfect greedy fitting procedure yielding non-optimal trees \citep{bertsimas2017optimal}.  Thus, randomization of trees through \texttt{B} \& \texttt{P} deals with unreliable optimization and performs early stopping.  Put together, these observations mean RF likely is pruning $\mathcal{T}$ in \eqref{eq:truetree}, when the DGP looks as such.
 
%\footnote{A simpler example is that of ridge regression: if there are more regressors than observations, then $\lambda \rightarrow 0$ leads to a multitude of solutions for $\beta$ and the non-identification of the predictive structure.} 
 
%\footnote{In Appendix \ref{sec:het}, I review a more standard case for Bagging based on presumed heteroscedascity.}  
 
The inspiration for the following argument comes from forecasting with non-linear time series models, in particular with the so-called Self-Exciting Threshold Autoregression (SETAR). A simple illustrative SETAR DGP is
\begin{align}\label{eq:setar}
y_{t+1}=\eta_t \phi_1 y_t  + (1-\eta_t) \phi_2 y_t +\epsilon_t, \quad \eta_t = I(y_t>0)
\end{align}
where $\epsilon_t$ is normally distributed. The forecasting problem consists in predicting $y_{t+h}$ for $h=1,...,H$ given information at time $t$. As it is clear from \eqref{eq:setar}, $y_{t+1}$ is needed to obtain the \textit{predictive function} for $y_{t+2}$ which is either $\phi_1$ or $\phi_2$. Alas, only an estimate \(\hat{y}_{t+1} = E\left( y_{t+1} \mid y_t \right)\) is available and, by properties of expectations, \(f\left( E\left( {y}_{t+1} \mid y_t \right) \right) \neq E\left( f(y_{t+1}) \mid y_t \right)\) if \(f\) is non-linear.  Hence, proceeding to iterate forward using $\hat{y}_{t+h}$'s as substitutes for ${y}_{t+h}$ at every step leads to a bias problem that only gets worse with the forecast horizon. If such an analogy were to be true for trees, this would mean that as the tree increase in depth, the more certain we can be that we are far from $\mathcal{T}(X_i)$, the optimal prediction function. I argue that it is the case.
 
%Hence, the interaction of non-linearity and recursive forecasting makes it impossible to obtain the unbiased forecast simply by inputting the last observed value $y_t$ and iterating forward (as is typical with linear autoregressive models). I will argue that an analogous phenomenon is at works for trees. 
 
%The bulk of problem in SETAR models is that $y_{t+1}$ is needed to obtained the predictive function for $y_{t+2}$ (the models switches between different set of coefficients according to $y_{t+1}$, hence 'self-exciting'). Alas, only $\hat{y}_{t+1}=f(y_t) = E(y_{t+1}|y_t)$ is available. 
 
%The proposition solutions in the literature are -- with a very familiar sound -- to either use parametric or non-parametric bootstrap to simulate the intractable expectation (\cite{clements1997setar}).\footnote{For a detailed discussion of the SETAR case and other non-linear time series models, see section 2.7 in \cite{NLTSA}.} I apply such thinking to the estimation of the true tree in \eqref{eq:truetree} with $\sigma_i = 1 \enskip \forall i$.

Following the time series analogy, the prediction for a particular $i$ can be obtained by a series of recursions. Define the cutting operator 
$$
\mathcal{C}(L; y,X,i) \equiv \mathcal{S}_i \left( \argmin _{k\in \mathcal{K}, \smallskip c \in {\rm I\!R}}\left[ \min\limits_{\mu_{1}} \sum \limits_{\{ i \in L| X_{i,k} \leq c  \}} \left(y_{i}-\mu_{1}\right)^{2} 
 + \min\limits_{\mu_{2}} \sum \limits_{\{ i \in L | X_{i,k} > c  \}} \left(y_{i}-\mu_{2}\right)^{2}\right] \right)
$$
where $\mathcal{S}_i$ extract the subset that includes $i$ out of the two produced by the splitting step. Inside the $\mathcal{S}_i$ operator is the traditional one-step tree problem. $\mathcal{K}$ is the set of potential features to operate the split at an optimized value $c$. $L$ is the leaf,  that is, the sample to split,  and is itself the result of previous cutting operations from steps $s-1$, $s-2$,  and so on.  To get the next finer subset that includes $i$, the operator is applied to the latest available subset: $L'=\mathcal{C}(L; y,X,i)$. The prediction for $i$ can be obtained by using $\mathcal{C}$ recursively starting from $L_0$ (the full data set) and taking the mean in the final $L$ chosen by some stopping rule. In other words, the true tree prediction in \eqref{eq:truetree} is $\mathcal{T}(X_i)=E\left(y_{i'} | i' \in \mathcal{C}^D(L_0; y,X,i)\right)$ where $D$ is the number of times the cutting operator must be applied to obtain the final subset in which $i$ resides. To obtain the true tree prediction -- the mean of observations in $i$'s "true" terminal node -- the sequence of $\mathcal{C}$'s must be perfect,  which is unlikely given the estimation error coming from both finite samples and myopic greedy optimization.   %Hence, consistency remains on safe ground: as the sample size grows large, estimation error vanishes and $\hat{\boldsymbol{S}} \rightarrow \boldsymbol{S}$ at each step. The finite sample story is, however, quite different. 
 
Using $\hat{y}_{t+1}$ instead of $y_{t+1}$ in SETAR and $\hat{L}$ instead of $L$ in a tree generate problems of the same nature.  At each step, the expected composition of $\hat{L}$ is indeed $L$.  However, just like the recursive forecasting problem, the expected \textit{terminal} subset is defined as an expectation over a recursion of non-linear operators. Using  $\hat{L}$ rather than the unobserved $L$ at each step does \textit{not} deliver the desired expectation. Intuitively, getting the right $k$ and $c$ out of many possible combinations is unlikely. These small errors are reflected in $\hat{L} \neq L$ which is taken \textit{as given} by the next step. Those errors eventually trickle down with absolutely no guarantee that they average out. In short, the direct CART procedure produces an unreliable estimate of a greedily constructed predictor $\mathcal{T}(X_i)$ because it takes as given at each step something that is not given, but estimated. Since $\mathcal{C}$ is a non-linear operator, this implies that the mean itself is not exempted from bias. 
 
% In other words, the expected prediction from the standard CART procedure is not $\mathcal{T}(X_i)$. 
 
 %\footnote{This notion can be formalized by defining the expectation in terms of indicator functions for each candidate observation. Each observation at each cutting step is expected to be classified in the right one of two groups.}
  
%$$
%\mathcal{C}(\boldsymbol{S}; y,X) \equiv \left\{ \boldsymbol{S} \bigg\rvert \min _{j, c}\left[\min _{\mu_{1}} \sum_{i \in \left\{\boldsymbol{S} | {X}_{j} \leq c\right\}}\left(y_{i}-\mu_{1}\right)^{2}+\min _{\mu_{2}} \sum_{i \in \left\{\boldsymbol{S} | X_{j}>c\right\}}\left(y_{i}-\mu_{2}\right)^{2}\right] \right\}
%$$
 
If the direct procedure cannot procure the right expected subset on which to take the average and predict, what will? The intuition for the answer, again, stems from the SETAR example. The proposed solution in the literature is -- with a distinctively familiar sound -- using bootstrap to simulate the intractable expectation (\cite{clements1997setar}).  $\hat{y}_{t+1}$ is augmented with a randomly drawn shock (from a parametric distribution or from those in the sample) and a forecast of $y_{t+2}$ is computed conditional on it. Then, the procedure is repeated for $B$ different shocks and the final forecast is the average of all predictions, which, by the non-linearity of $f()$, can make it a very different quantity from $f(\hat{y}_{t+1})$. A forecaster will naturally be interested in more than $y_{t+2}$. This procedure can be adapted by replacing the draw of a single shock by a series of them that will be used as the model is simulated forward. The prediction at step $H$ is an average of forecasts at the end of each $b$ randomly generated sequence.
 
%\footnote{For a discussion of the SETAR case and other non-linear time series models, see section 2.7 in \cite{NLTSA}.}
 
Analogously, a natural approach is to simulate the distribution of $L$ entering a next splitting step is to bootstrap the sample of the previous step, run $\mathcal{C}$ a total of $B$ times, apply $\mathcal{C}$ in the next step and finally take the average of these $B$ bootstrapped trees predictions. For a deeper tree, the growing process continues on the bootstrapped sample and the average is taken once the terminal condition is reached. 
 
Coming to the original question: if RF is pruning something, what is it? I conjecture it is pruning $\mathcal{T}$ in \eqref{eq:truetree}. Unlike the implicit early-stopping property explained in section \ref{sec:rf_miracle}, this statement cannot be supported or refuted by the simulations presented in section \ref{sec:simuls}. However, in \cite{SGT}, it is shown that under a "true tree" DGP, the performance of RF and a version of CART with a low learning rate coincides. The latter can be linked to fitting the true tree optimally via an (extremely) high-dimensional LASSO problem.  
 
Finally,  when the DGP looks nothing like a tree, it is understood that RF supplants CART anyway since it smooths CART's hard decision boundary \citep{wyner2017explaining}.  Rather, the arguments in this section (and the paper, more generally) focus on why RF is better than a single tree even when the DGP itself is a tree.  In those cases, smoothing can hardly be the sole explanation of RF success (as exemplified by using a tree DGP in the  top-left quadrant of Figure \ref{simul_nl2}).

\section{Comparison with Current Justifications of RF's Success}\label{sec:comp}
 
In this section, I compare this paper's proposition to the two main alternatives currently available.    
 
\subsection{Not Your Average Model Averaging}\label{sec:modavg}
 
%in short, because it also bring bias down
%but does so by fixing the optimization algo rather than increasing capacity
This paper is certainly not the first to attempt explaining RF's success. The original explanation comes from \cite{breiman2001} showing that RF generalization error is bounded.  The textbook intuition is that overfitting trees have very low bias and ensembling them reduces variance without increasing bias \citep{ESL}.  The previous section notes that RF's behavior is not only good in terms of bias-variance trade-off (as the generalization bound suggests),  it is optimal.  It also notes that ensembling diversified zero-bias linear regressions does not work as well as RF,  even in linear DGP environments --- thus,  the complexity of base learners still requires tuning \citep{elliott2013complete}. Thereby,  trees are special beyond their non-linearity, and RF's success cannot solely be attributed to the averaging of low bias and high variance base learners.
 
%\textbf{\color{red} NEW} 
 
%At first sight, the phenomenon described here may seem like nothing new: RF successfully controls overfitting by approximating more resampling by model averaging. 
 
Well before the inception of ensembling ideas in machine learning,  there was (and there still is) a wide literature in statistics and econometrics about model averaging and forecast combinations.  The latter is known to provide a sort of regularization that can, in some special cases, be equivalent to more traditional shrinkage estimators \citep{elliott2013complete,lejeune2020implicit}. What is new in RF is that unlike averaging a kitchen-sink OLS regression (for instance), a greedy algorithm makes the structure estimated before $s^*$ immune to what happens in the overfitting zone. In contrast, schemes like that of \cite{eliasz2004optimal} or those discussed in  \cite{rapach2013forecasting} imply directly or indirectly tuning the number of regressors in the base learner linear models. This means that including too many of them could damage the overall model’s performance.  Figure \ref{modavg} makes this distinction clear: OLS, while experiencing a reasonable performance renaissance thanks to double descent, could still strongly benefit from tuning the number of included regressors.  Of course, all of this is analogous to tuning $\lambda$ in a ridge regression, and the resulting $R^2_{\text{test}}$ is usually in the neighborhood of $R^2_{\text{train}}$. This is why using a (global) linear model to think about \texttt{mtry}'s effect  -- while it may yield interesting insights \citep{ESL,mentch2019randomization} --  provides an incomplete answer that fails to capture one of RF's most salient regularities: $R^2_{\text{train}}>R^2_{\text{test}}$.  Hence, \texttt{B} \& \texttt{P} are not the source of the “paradox” \textit{per se}: they must be paired with a greedy algorithm which can generate sufficient inner randomization. 
 
%cite diebold ijf
 
Conversely, the ideas presented above could help in understanding why forecast combinations work so well, which unlike confirming their multiple successes, is still an ongoing venture \citep{timmermann2006forecast}.  It is plausible that individual forecasters construct their predictive rule in an inductive recursive fashion. That is, human-based economic forecasting has plausibly more to do with a decision tree or a stepwise regression,  than with OLS or Lasso.  Indeed,  it is arguably much easier to learn in a greedy fashion (either for a human and a computer) than to solve a complex multivariate problem directly for its global solution. Thus, assuming underlying forecasts are constructed as such, the average will behave in a very distinctive way if those are overfitting. As argued in \cite{hellwig2018overfitting} for the survey of IMF forecasters, the latter assertion is very likely true. As a result, the discussion above provides yet another explanation for the success of forecast combinations (especially the simple average scheme): significant inner randomization combined with recursively constructed overfitting forecasts provides implicit pruning. This is not a replacement but rather a complement to traditional explanations (usually for linear models) that link the effects of model averaging to traditional shrinkage estimators \citep{ESL,rapach2013forecasting}.   %\footnote{For instance, \cite{} discuss the link for RF itself, \cite{} discuss it for the case of forecasting stock returns with averages of linear models.} 
 
%\footnote{This is especially true of the resilience of equal weights \citep{egalasso}. }
 
\subsection{Double Descent vs. a Single Monotonic Descent in Random Forest}\label{sec:nstdd}
 
%MERGE P1 and P2
 
It is worth contrasting this paper's explanation with recent "interpolating regime" and "double descent" ideas proposed to explain the success of deep learning \citep{belkin2019reconciling,belkin2019does,hastie2019surprises,bartlett2020benign,kobak2020optimal}.  In a regression context, the interpolating regime is entered whenever one fits an algorithm of ever-increasing complexity past the complexity level delivering $R^2_{\text{train}}=1$.  The association with "interpolation" as traditionally defined in mathematics comes from training data points being effectively interpolated by the fitted function $\hat{f}$ when $R^2_{\text{train}}= 1$.  The double descent is the astonishing observation that for large-scale deep neural networks (DNN), the out-of-sample performance starts to increase past the point where $R^2_{\text{train}}=1$. Preceded by the typical U-shaped empirical risk curve implied the classical bias--variance trade-off before $R^2_{\text{train}}=1$, this makes it for a "double descent" --- the first starting from $R^2_{\text{train}}=0$ and the second from $R^2_{\text{train}}=1$.

%However, their construction mistakenly associates the number of trees to be increasing complexity (as in Boosting) whereas it is explicit increased averaging/regularization in RF. 
 
%Thus, there is no double descent in RF, but rather a single \textit{monotonic} descent. Section \ref{sec:nstdd} elaborates on this matter.  In an influential paper, \cite{belkin2019reconciling} argue that Random Forest (RF), along with deep neural networks, exhibits a phenomenon now known as "double descent". 
 
\cite{belkin2019reconciling} state that the phenomenon is also present in RF.   Its occurrence for tree ensembles is an artifact of a strangely constituted notion of function capacity. \cite{belkin2019reconciling} define it as increased tree depth until interpolation occurs, and augmenting the number of trees past that threshold.  While it is officially correct, the construction is misleading since the number of trees is not a tuning parameter \citep{scornet2017tuning}, unlike the number of neurons and layers in a neural network.  As shown in \cite{ESL}, one needs to set $B$ large enough so that the prediction function stabilizes,  and the cost of larger $B$'s is solely computational.  
 
Unsurprisingly,  in \cite{belkin2019reconciling}'s example,  the MSE starts to decrease sharply with $B$. The problem is, it does not do so because of increased functional complexity (like increasing the number of layers/neurons in a DNN, or regressors in a very large Ridge regression), but rather because \textit{explicit} regularization (i.e., ensembling) has increased. More trees in RF is akin to simulating enough draws from a distribution of initial conditions so that the average model stabilizes, so it is not surprising to see that an ever-increasing number of trees brings down MSE until it reaches a plateau \citep{ESL} -- it is the law of large numbers.  RF may have more parameters than a single tree, but the different trees are never used together to fit the same $y_{\text{train}}$,  like in Boosting.  As a result, averaging trees procuring $R^2_{\text{train}}<1$ in isolation will \textit{not} help the forest $R^2_{\text{train}}$ climb to 1 --- i.e.,  in certain cases, they can even decrease function capacity and shrink $R^2_{\text{train}}$.  This is so because the additional parameters directly serve the purpose of model averaging (which can, in linear models, have similar effects to Ridge regularization,  \citealt{elliott2013complete}).  %Thus, 
 
Finally, \cite{belkin2019reconciling}'s U-shaped curve before $R^2_{\text{train}}=1$ is expected because they use a single tree (similar curves are reported for plain CART in Figure \ref{simul_nl2}), which is not a forest.  Thus, with an appropriately defined notion of base learner capacity -- like underlying trees \textit{depth},  which will,  always and everywhere,  increase $R^2_{\text{train}}$ -- there is no such thing as double descent of empirical risk for RF.  Rather, as put forward in this paper, we get something better: a single \textit{monotonic} descent.
 
%\footnote{Moreover, constituent trees, the true modulator function capacity, cannot have more parameters than observations: once each observation is in its own cell, optimization is over.} 
 
\begin{figure}[h!] %{0.5\textwidth}
\vspace*{-0.2cm}
  \begin{center}
    \includegraphics[trim={0.25cm 0.75cm 0cm 0cm},clip, width=0.5\textwidth]{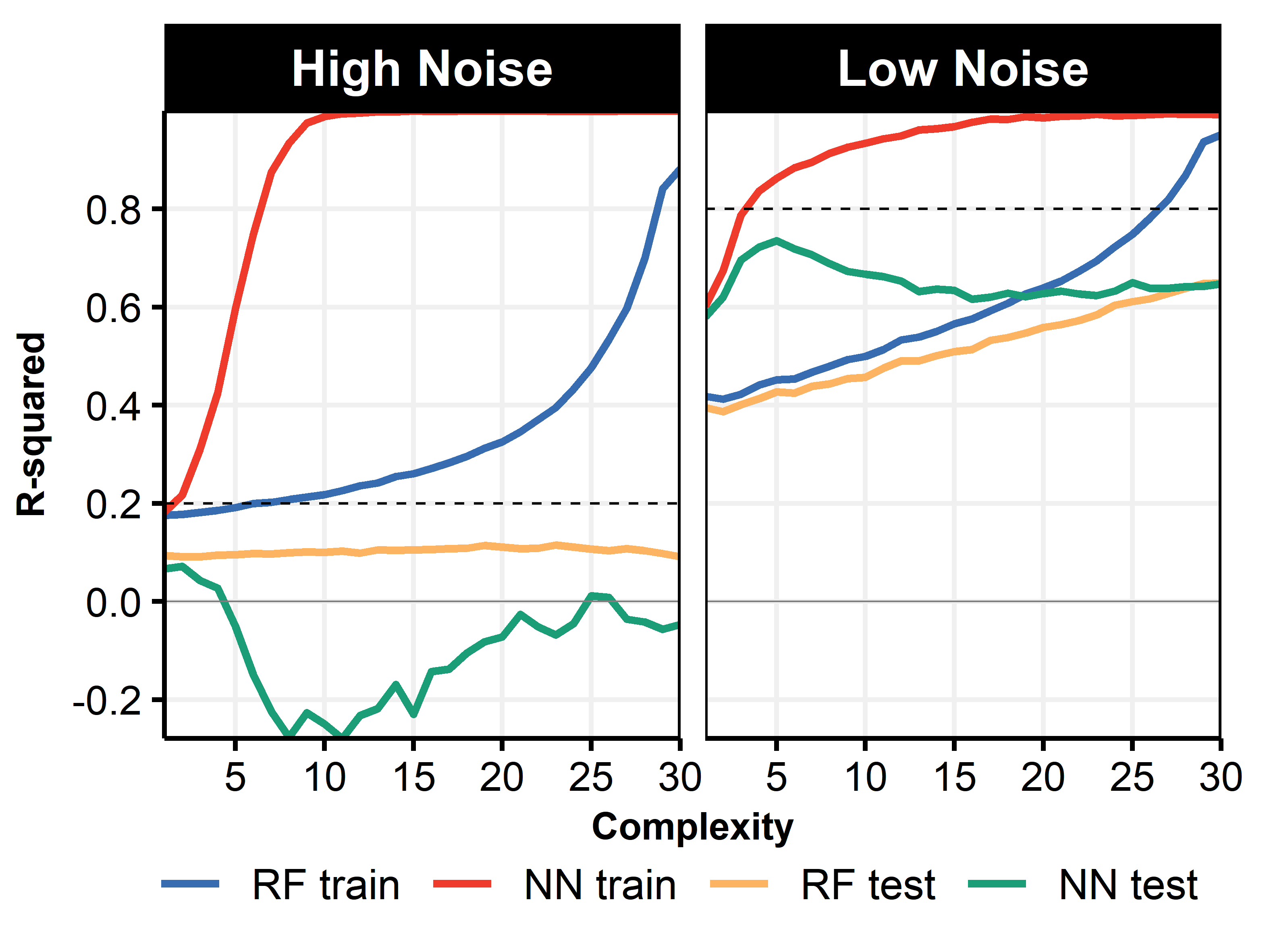}
  \vspace{-0.1cm}  
  \caption{\footnotesize  Dashed lines are true $R^2$, i.e., the best attainable $R^2$ out-of-sample given the variance of the true error.  DGP is Friedman 1 \citep{MARS} simulated using the \href{https://www.rdocumentation.org/packages/mlbench/versions/2.1-3/topics/mlbench.friedman1}{\texttt{mlbench} package.}.  Precisely,  inputs are 10 independent variables uniformly distributed on the interval \([0,1]\), only 5 out of these 10 enter the DGP so that $y_i = 10 \sin(\pi x_{1,i} x_{2,i}) + 20 (x_{3,i} - 0.5)^2 + 10 x_{4,i}+ 5 x_{5,i} + \epsilon_i$ with $\epsilon_i$ is normal noise (so it is optimal for both algorithms to minimize the squared loss).  The $x$-axis is an index of complexity/depth. For RF, it is a decreasing minimal size node from 200 to 1 in 30 steps, and for NN, an increasing number of layers from 1 to 30. The NN is 50 neurons wide and RF's $\texttt{mtry}=\sfrac{1}{3}$.  Other NN details are in Appendix \ref{sec:nndetails}.}
  \label{nn_vs_rf}
    \end{center}
  \vspace*{-0.4cm}  
\end{figure}

\nocite{geiger2020scaling,d2020double}

Figure \ref{nn_vs_rf} compares NN and RF. In the high noise environment, even though NN out-of-sample performance can experience a revival in the overly parametrized regime, it remains far inferior to that of RF, which remains \textit{constant} as its $R^2_{\text{train}}$ goes way beyond the true $R^2$. Strikingly, while NN's depth needs to be tuned in each environment, the best tree depth is always the deepest one whether $y$ is mostly noise or signal.\footnote{Some choices needed to be made in how to put in correspondence model capacity of a RF and a NN.  First, I chose two HPs in both models that have a clear direct effect of increasing model capacity as showcased by training set performance in both cases. The choice of minimal node size for RF is a direct implication from the discussion about RF performs implicit optimal early stopping.  Manual early stopping can be done with minimal node size, so it is a natural choice.   "Number of layers” is chosen for the NN part of the graph because it clearly increases the number of parameters and the capacity of the model as showcased by the red curves in Figure  \ref{nn_vs_rf}  -- and more than network width in this case.} Thus, when it comes to its crucial hyperparameter guiding function capacity (see blue line), there is no tuning problem in RF.  Seen differently,  if "nature" is choosing the noise level,  then opting for the deepest trees is the modeler's dominant strategy.  Of course, this is an implication of the earlier discussion: explicit pruning or no pruning at all yield nearly identical performance.
 
It is not entirely surprising that RF behaves differently from NN, with their respective estimation being carried very differently. For NN (or Ridge) the number of parameters is fixed during the estimation/optimization whereas it is constantly evolving for a tree. That is, gradient descent methods typically estimate a model of fixed complexity globally. If that complexity level is too great, the model will fail to generalize. In contrast, for greedy methods, estimation/optimization and model building go hand in hand, and, importantly, different parameters are estimated in different stages within "transitory models" of varying complexity. Thus, it is no surprise that double descent does not transfer to RF, and that self-pruning is not expected from a NN.
 
Finally,  \cite{wyner2017explaining} also argue that interpolation may be the key for Boosted Trees and RF success because local fitting of dissident data points prevents harming the overall prediction function $\hat{f}$. But it is unclear as to why RF is so proficient at it, why "locality" emerges in the first place, and why estimation variance does not spread. The current paper makes exactly clear how the (greedy) construction of RF guarantees that overfitting washes away out-of-sample.
 
%\section{Experiments}
\section{Simulations}\label{sec:simuls}
 
Simulations are carried to display quantitatively the insights presented in the previous section. Namely, I want to display that \textbf{(i)} ideal population sampling of greedy algorithms performs pruning/early stopping, \textbf{(ii)} RF very closely approximates it for trees, and \textbf{(iii)} the property also extends to altered versions of Boosting and MARS. 
 
\subsection{Setup}
 
I consider 3 versions of 3 algorithms on 5 DGPs. The 3 models are a single regression tree (CART), Stochastic Gradient Boosting (with tree base learners) and MARS.  The first two versions of each model are obvious.  First, I include the plain model and second, a bootstrapped and perturbed ensemble of it, as described earlier.  For the CART base learners,  this corresponds to RF.  For MARS and Boosting,  this is MARSquake and Booging,  as introduced in section \ref{sec:expend}.
 
The third version of each model, "Population Sampling" aims at displaying what results look like under the ideal case of perfect randomization.  Subsampling is replaced by sampling $B$ non-overlapping subsets of $N$ observation from a population of $B \times N$ observations. This version help discern which algorithm generates enough inner randomization to get close to that desirable upper bound.  The idea is that, if there were no data limitations, non-overlapping samples of the same DGP  is the best way to average overfitted noise out.  \texttt{B}+\texttt{P} is inevitably imperfect because the same observations are recycled in different “bags” allowing for the possibility that bits of overfitted $E(y_i|x_i)$ do not average out to zero. This does not happen with population sampling.  Thus, it makes empirically operational the ideal experiment of the Perfectly Random Forest utilized in \ref{sec:prf},  and that is why for any greedy algorithm under consideration, the purple line will be perfectly monotically increasing. 
 
The five DGPs are a Tree, Friedman 1, 2 and 3  \citep{MARS,breiman1996bagging} as well as a linear model.  The true tree DGP is generated using a CART algorithm's prediction function as a "new" conditional mean function from which to simulate. The "true" minimal node size being used is 40 (10\% of the training set). Friedman DGP are obtained from the package \texttt{tgp} and typically generate a data set with 5 useful regressors (used in the true DGP) and 5 useless ones.  Further details about how the data is simulated is in Appendix \ref{sec:simdet}.

%\footnote{Appendix \ref{sec:addi_graph} reports results when $\nu=0.01$ and $\nu=0.3$.}

%Figures \ref{simul_nl2} and \ref{simul_nl1}. 
 
In terms of standard hyperparameters, Boosting has the shrinkage parameter $\nu=0.1$, the fraction of randomly selected observations to build trees at each step is 0.5, and the interaction depth of those trees is 3. Of course, while those are fixed for all simulations, we will want to tune them once we get to real data. However, here, the point is rather to study the hold-out sample performance of each model as its depth increase, and compare that across the 3 versions. MARS has the interaction degree set to 3,  meaning that a basis function $h_s$ can at most be the interaction of 3 hinge functions based on (possibly) 3 different variables.  RF is used with a rather high \texttt{mtry} of $\sfrac{9}{10}$ so to be better visually in sync with plain CART at a given depth.\footnote{For completeness, results when using \texttt{mtry}= $\sfrac{1}{2}$ for both plain CART and RF are reported in Figure \ref{simul_nl2_mtry50}. It is clear that in the high signal-to-noise ratio environment, the milder perturbation of \texttt{mtry}=$\sfrac{9}{10}$ is preferable.} The subsampling rate is $\sfrac{2}{3}$ for all bagged models. 
% --- e.g., this is the case for Friedman's DGPs which have 5 useful regressors and 5 useless ones.

\begin{figure}[t!]
\begin{center} 
\hspace*{-0.4cm}\includegraphics[scale=.6]{{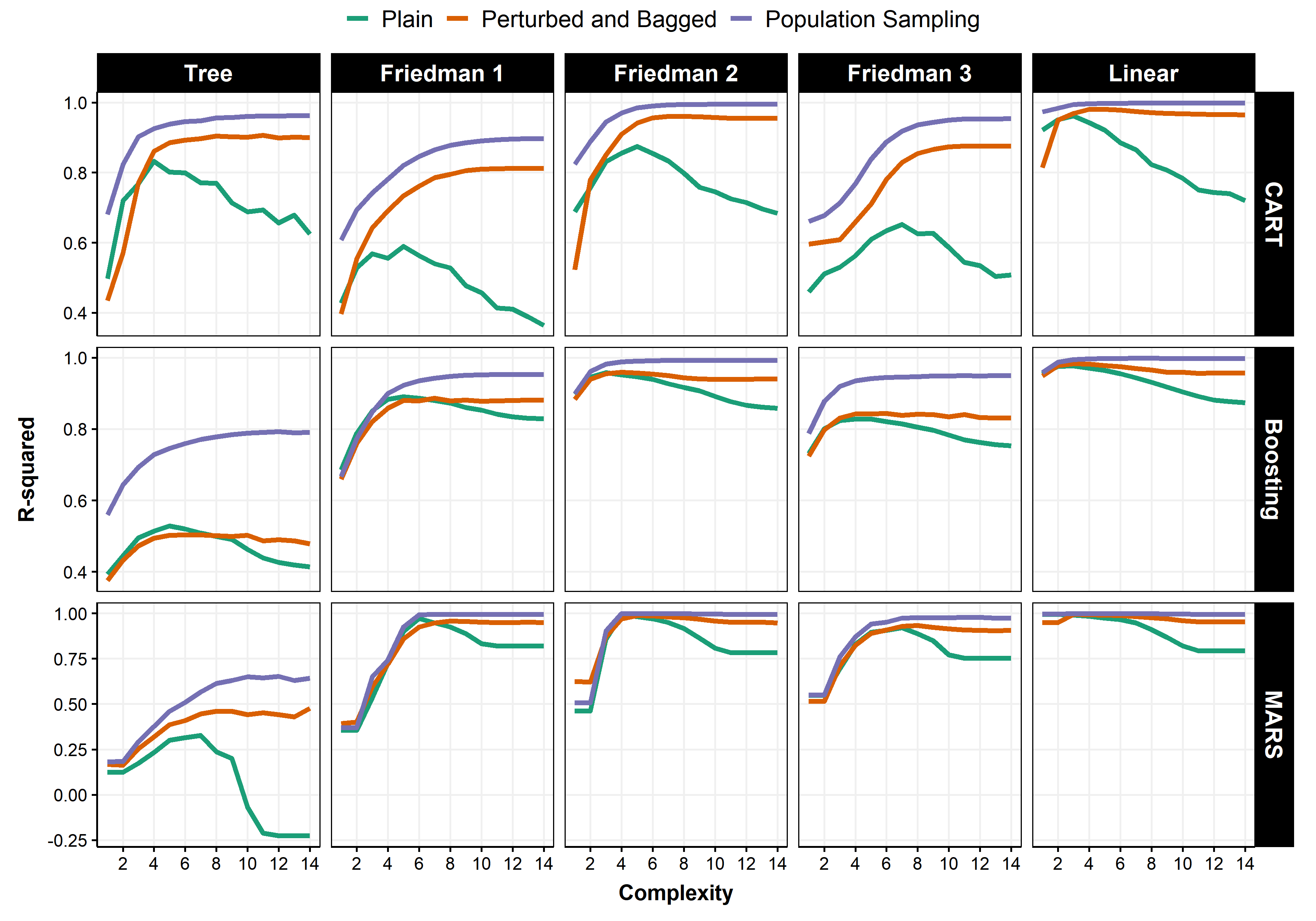}}
\vspace*{-0.4cm}  
\caption{\scriptsize This plots the average hold-out sample $R^2$ between the prediction \textbf{and the true conditional mean} for 30 simulations. The level of noise is calibrated so the SNR is 4. Column facets are DGPs and row facets are base learners. The $x$-axis is an index of depth of the greedy model. For CART, it is a decreasing minimal size node $\in 1.4^{\{16,..,2\}}$, for Boosting, an increasing number of steps $\in 1.5^{\{4,..,18\}}$ and for MARS, it is an increasing number of included terms $\in 1.4^{\{2,..,16\}}$. Both training and test sets have $N=400$.}
\label{simul_nl2}
\end{center}
\vspace{-0.25cm}  
\end{figure}
 
%Additionally, it corresponds directly to the perfect randomization scenario necessary for perfect self-pruning to take place.
\nocite{earthpackage}
 
\subsection{Results}
 
Figures \ref{simul_nl2}  and \ref{simul_nl1} report the median $R^2$ between hold-out sample predictions and the true conditional mean for 30 simulations. Columns are DGPs and rows are models. The $x$-axis is an increasing index of complexity for each greedy model. Overfitting should manifest itself by a decreasing $R^2$ past a certain depth. I consider two levels of noise, one that corresponds to a SNR ratio of 4 (Figure \ref{simul_nl2}) and one of 1 (Figure \ref{simul_nl1} in Appendix). What does section \ref{sec:rf_miracle} imply for the curves in those two Figures? First, the population sampling versions (purple line) should be weakly increasing since they perform implicit "perfect" early stopping. Second, the \texttt{B} \& \texttt{P} versions (orange), should be parallel to those provided the underlying greedy model is generating enough inner randomization. Third, the value of the orange line at the point of maximal depth should be as high or higher than the maximal value of the green curve (i.e, the plain version's ex-post optimal stopping point). 
 
When it comes to CART, those three properties are verified exactly. For any DGP, and both the population sampling and the \texttt{B} \& \texttt{P} versions, increasing the complexity of the model by shrinking the minimal node size does \textit{not} lead to a performance metric that eventually decrease. The striking parallelism of the purple and orange lines is due to trees generating enough inner randomization with \texttt{B} \& \texttt{P} so it performs self-pruning at a level comparable to that of the ideal experiment.  Note that the overall level of purple line is mechanically expected to be at least above the orange one for a fixed depth: the former uses more data points which also helps at reducing estimation error.  Depending on the DGP, the plain version follows a typical bias-variance trajectory: it generally follows the \texttt{B} \& \texttt{P} one for some time before detaching itself from it past its ex-post optimal $s^*$. This early parallelism of the  green and orange lines corroborate the idea that \texttt{B} \& \texttt{P} CART (i.e.,  RF) performs implicit pruning.
 
Looking at Boosting and MARS, we again see that the population sampling line is weakly increasing in the respective depth of both models. If the \texttt{B} \& \texttt{P} version fails to match this ideal shape, it is because the current specification cannot generate enough inner randomization. Figure \ref{simul_nl2} shows unequivocally encouraging results for both Boosting and MARS. For all DGPs, a clear pattern is observed: the \texttt{B} \& \texttt{P} version's performance increases until it approximately reaches the optimal point (as can be ex-post determined by the hump in the green line) and then \textit{remains at that level}, even if the base learners (one example being the 'Plain' version) are clearly suffering from overfitting. Under those conditions, it is fair to say that the enviable RF property is transferable to Boosting, and in a more pronounced fashion, MARS. When the noise level increase as depicted in Figure \ref{simul_nl1}, we observed the same -- albeit marginally less successful -- phenomenon. Indeed, in those harder conditions, there is a small gap between ideal randomization and the one generated by Booging and MARSquake. However, the decrease in performance following the optimal depth is orders of magnitude smaller than what is observed for the plain version.
 
\section{Empirics}\label{sec:empirics}
 
\nocite{rasmussen1997evaluation} 
 
Applying \texttt{B} \& \texttt{P} to optimally-stopped Boosting and MARS is nothing new in itself. For instance, \cite{rasmussen1997evaluation} report that Bagged MARS supplant MARS on many data sets. This section has a more subtle aim than crowning the winner of a models' horse race. Rather than focusing on improving the tuned/pruned model which is already believed to be optimal, \textit{Booging} and \textit{MARSquake} bag and perturb completely overfitting based learners, which, as we will see, perform very poorly by themselves.  Their performance will be compared to versions of Boosting and MARS where the optimal stopping point has been tuned by CV. The goal is to verify that in many instances, Booging and MARSquake provide similar predictive power to that of tuned models. Since CV's circumstantial imperfections are vastly documented \citep{krstajic2014cross, bergmeir2018note}, it is not unrealistic to expect the \texttt{B} \& \texttt{P} versions to sometimes outperform their tuned counterparts. %In sum, this small application shows that the equivalence championed in previous chapters holds with real data and thus provides data scientists with a fruitful alternative to consider when building models.

\subsection{Setup}
 
Most data sets are standard with a few additions which are thought to be of interest. For instance, many of the standard regression data sets have a limited number of features with respect to the number of observations. A less standard inclusion like \textit{NBA Salary} has 483 observations and 26 features. \textit{Crime Florida} pushes it much further with a total of 98 features and 90 observations. Those data sets are interesting because avoiding CV could generate larger payoffs in higher-dimensional setups. Still in the high-dimensional realm, but with the additional complication of non-\textit{iid} data, are the 6 US macroeconomic data sets based on \cite{mccracken2020fred}.\footnote{Bagging has received attention of its own in the macroeconomic forecasting literature \citep{inoue2008useful,hillebrand2010benefits,hillebrand2020bagging}. However, nearly all studies consider the more common problem of variable selection via hard-thresholding rules -- like t-tests \citep{lee2020bootstrap}. Those strategies are akin to what discussed in section \ref{sec:modavg}, and cannot (and do not) strive for automatic pruning. Nevertheless, the motivation for using Bagging in their context is very close to what described for trees in section \ref{sec:setar}.} Moreover, traditional CV can be overoptimistic in a time series context and avoiding it could help \citep{bergmeir2018note}. The 3 macroeconomic variables are quarterly GDP growth, unemployment change and inflation. I consider predicting those variables at an horizon of 1 quarter ($h=1$) and 2 quarters ($h=2$). Further information on data sets is gathered in Table \ref{tab_datasets}. I do a 70-30 training-test split for all data sets. 
 
Beyond Boosting, MARS, and their novel variants, I include a few benchmark models. Those include LASSO, RF with default tuning parameters (\texttt{mtry}=$\sfrac{1}{3}$), a cost-complexity pruned regression tree, and two different neural networks.  The first NN is shallow (2 layers of 32 and 16 neurons) and is inspired from \cite{kellyml}. Such an architecture has provided reasonable performance on Canadian \citep{rapportMFQ} and UK macroeconomic data \cite{GCMS}. The second is a deep NN (DNN, with 10 layers of 100 neurons) following the recommendations of \cite{olson2018modern} for small data sets.  Many additional NNs details are in Appendix \ref{sec:nndetails}. For macro data sets, benchmarks additionally include an autoregressive model of order 2 (AR) and a factor-augmented regression with 2 lags (FA-AR) which are widely known to be hard to beat \citep{stock1999forecasting,KLS2019,GCLSS2018}.  %Remaining details needed to replicate results are in Appendix \ref{sec:empdet}.

%\subsection{Simulations Details}\label{sec:simdet}
 
For all data sets, I keep 70\% of observations for training (and optimizing hyperparameters if needed) and the remaining 30\% to evaluate performance. For cross-sectional data sets, those observations are chosen randomly. For time series applications, I keep the observations that consist of the first 70\% in the sample as the training set. The test set starts before the 2001 recession and ends in 2014, which conveniently includes two recessions. Lastly, a seldomly binding outlier filter is implemented. Every prediction that is larger than twice the maximal absolute difference (in the training sample) with respect to the mean is replaced by the RF prediction (which is immune to outliers since it cannot extrapolate). This last addition is particularly helpful to prevent wildly negative $R^2_{\text{test}}$ for non-tuned plain MARS and (less frequently) Boosting.
 
The $X$ matrix for the macroeconomic data sets is based on \cite{MRF}'s recommendations for ML algorithms when applied to macro data, which is itself a twist (for statistical efficiency and  less computational demand) on well-accepted time series transformations (to achieve stationarity) as detailed in \cite{mccracken2020fred}.  Each data set has 212 observations and around 600 predictors. The number of features varies across macro data sets because a mild screening rule was implemented ex-ante, the latter helping to decrease computing time.
 
%\footnote{\cite{GCLSS2018} further study optimal data transformations for machine macroeconomic forecasting for many series and algorithms.} 
 
\subsection{Results}
 
\begin{figure}[htbp]
%\centering
\captionsetup[subfigure]{justification=centering}
\setlength{\lineskip}{3.2ex}% increase spacin
  \begin{subfigure}[b]{0.47\linewidth}
\includegraphics[trim={1cm 1.1cm 0.25cm 1.1cm},clip,scale=.35]{{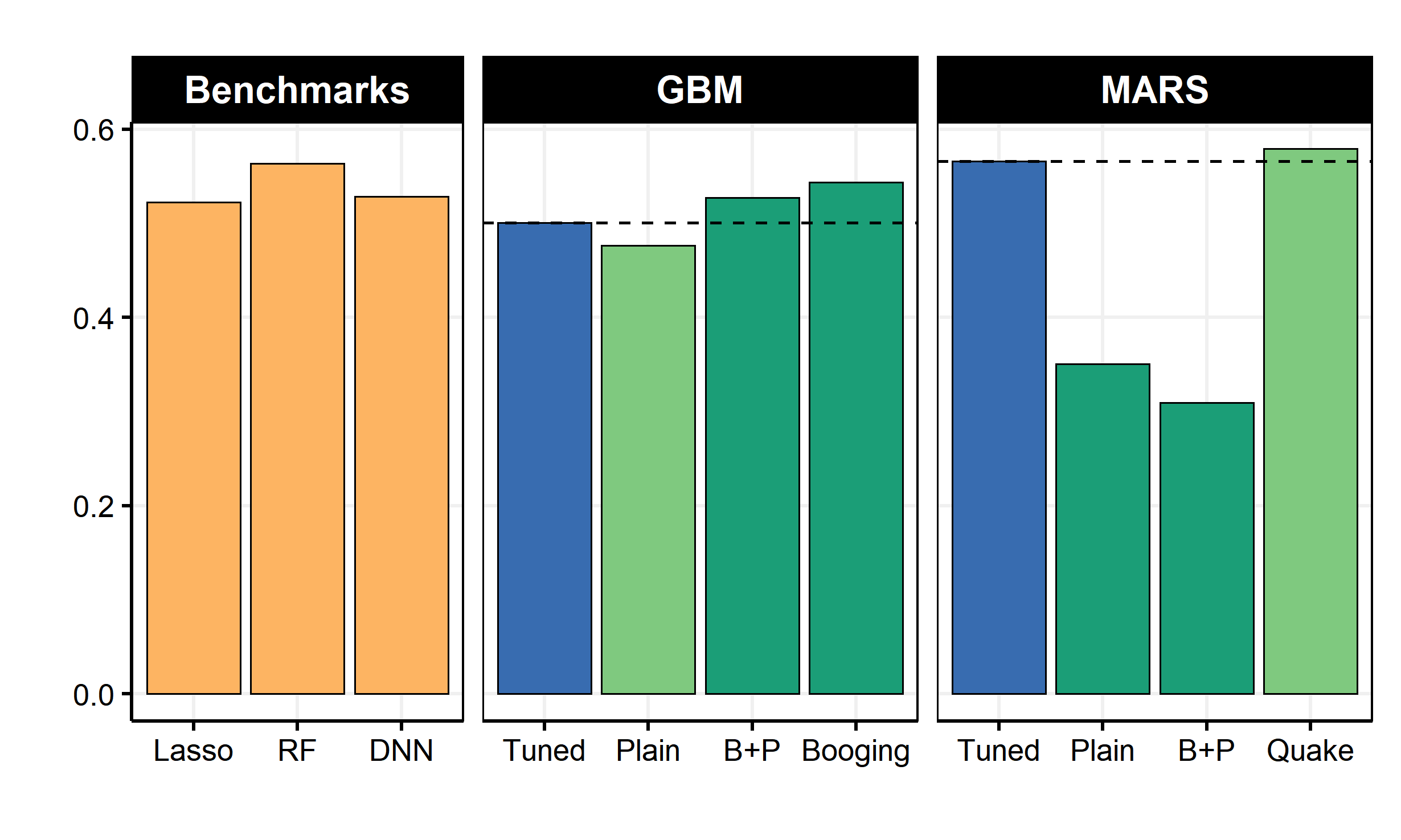}}
\caption{\textit{Abalone}}  
\label{IP_EOTB_detail}
  \end{subfigure}
  \begin{subfigure}[b]{0.47\linewidth}
\includegraphics[trim={1cm 1.1cm 0.25cm 1.1cm},clip,scale=.35]{{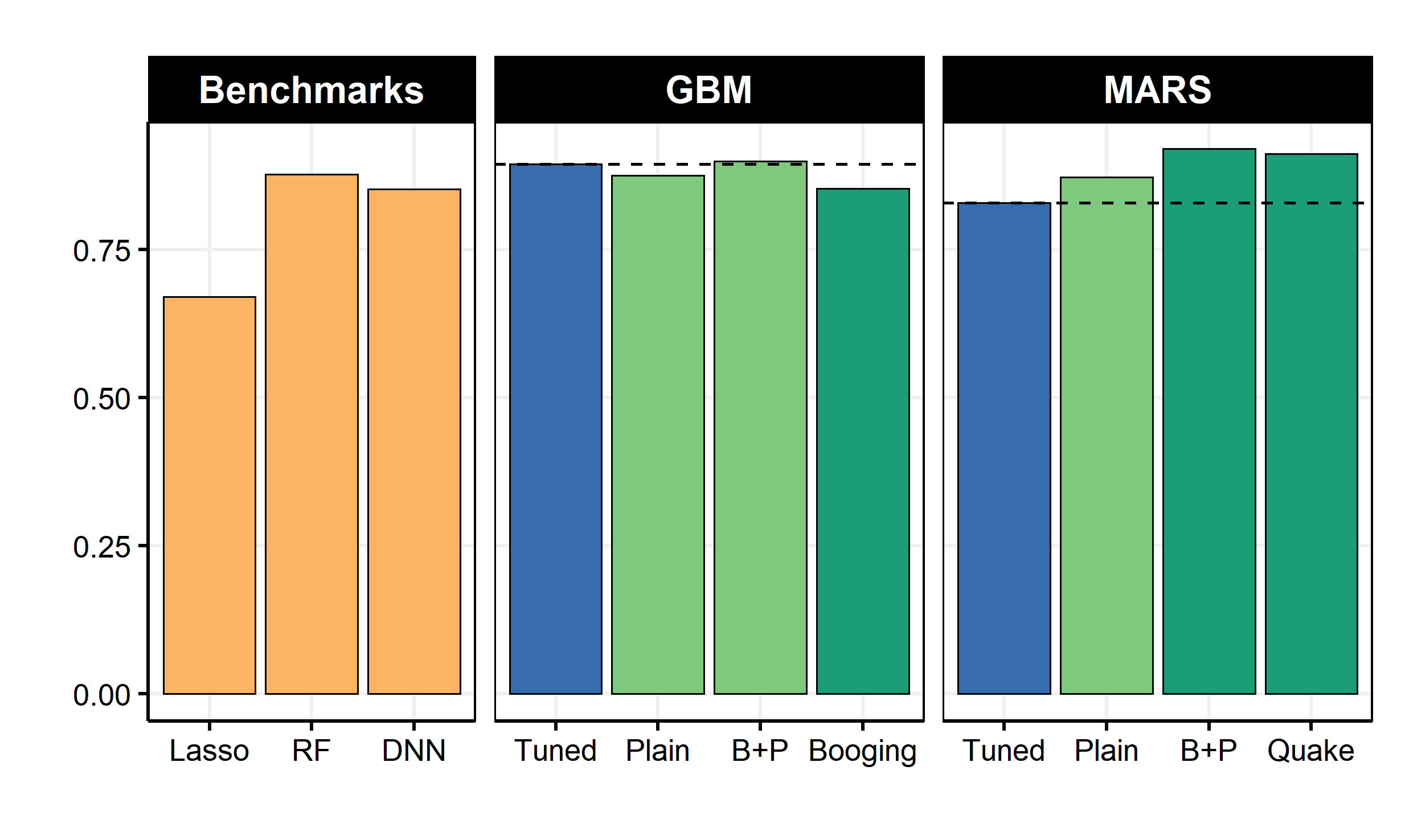}}
\caption{\textit{Boston Housing}}  
\label{UR_EOTB_detail}
    \end{subfigure}
  \begin{subfigure}[b]{0.47\linewidth}
\includegraphics[trim={1cm 1.1cm 0.25cm 1.1cm},clip,scale=.35]{{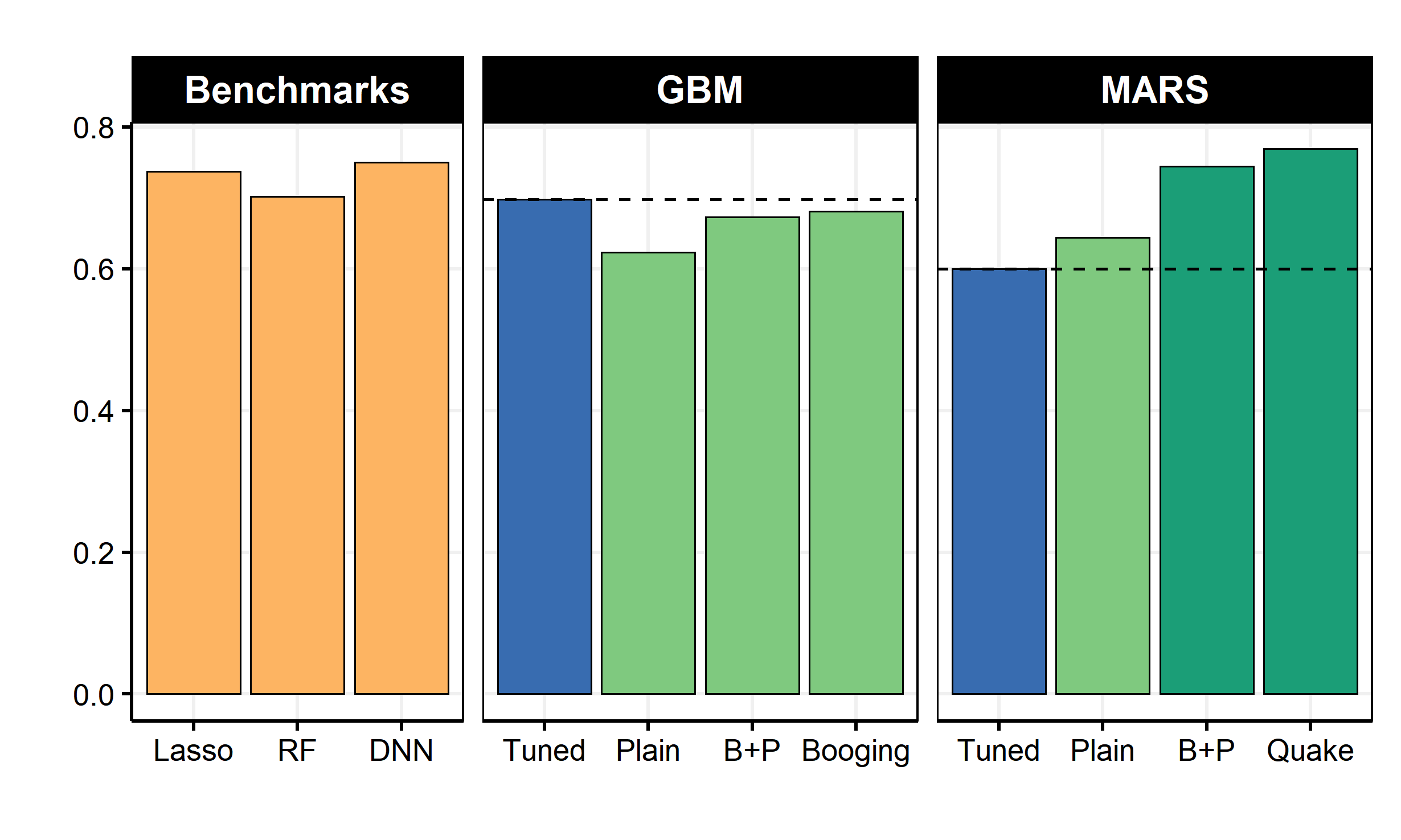}}
\caption{\textit{Crime Florida}}  
\label{SPREAD_EOTB_detail}
     \end{subfigure}
  \begin{subfigure}[b]{0.47\linewidth}
\includegraphics[trim={1cm 1.1cm 0.25cm 1.1cm},clip,scale=.35]{{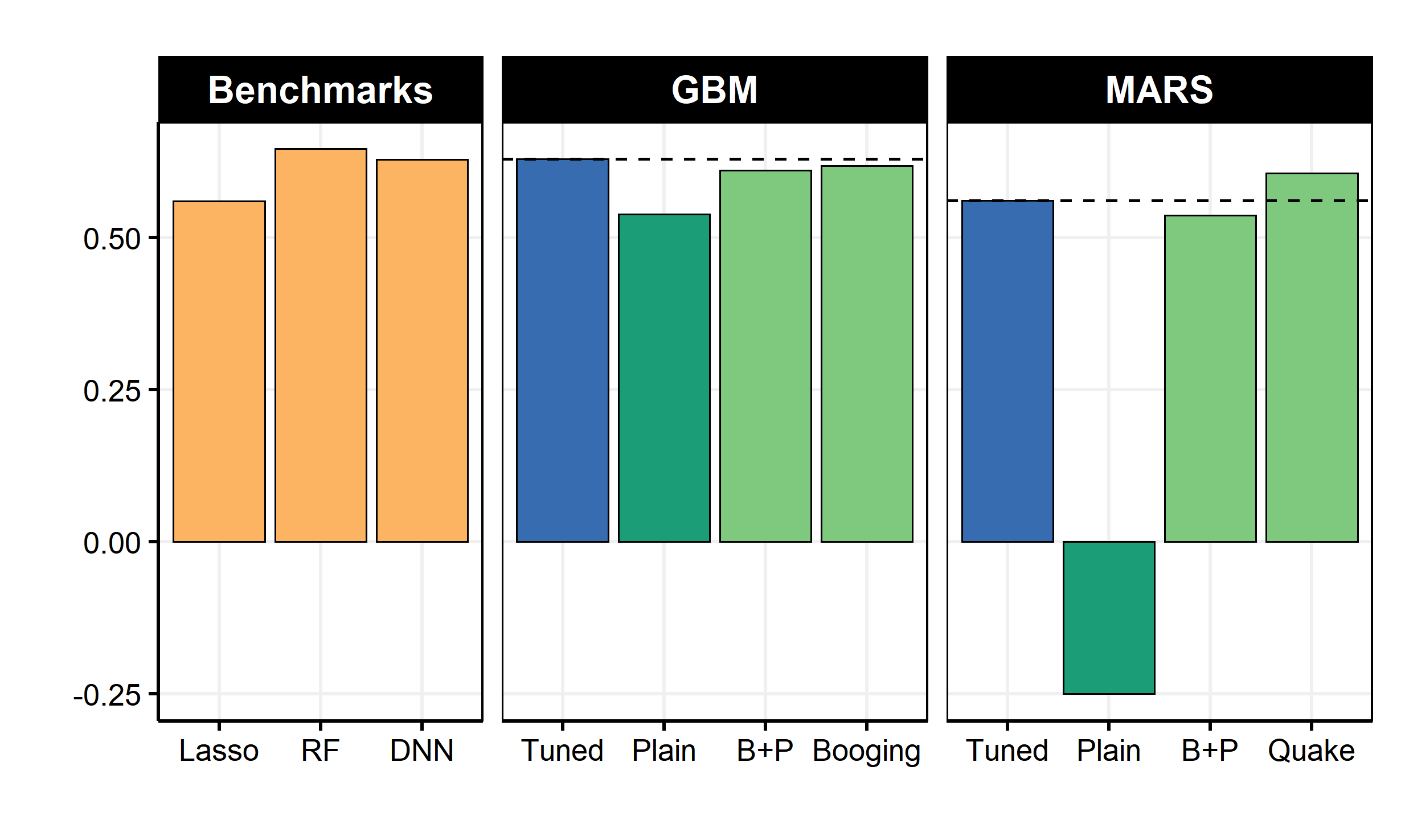}}
\caption{\textit{Fish Toxicity}}  
\label{INF_EOTB_detail}
      \end{subfigure}
  \begin{subfigure}[b]{0.47\linewidth}
\includegraphics[trim={1cm 1.1cm 0.25cm 1.1cm},clip,scale=.35]{{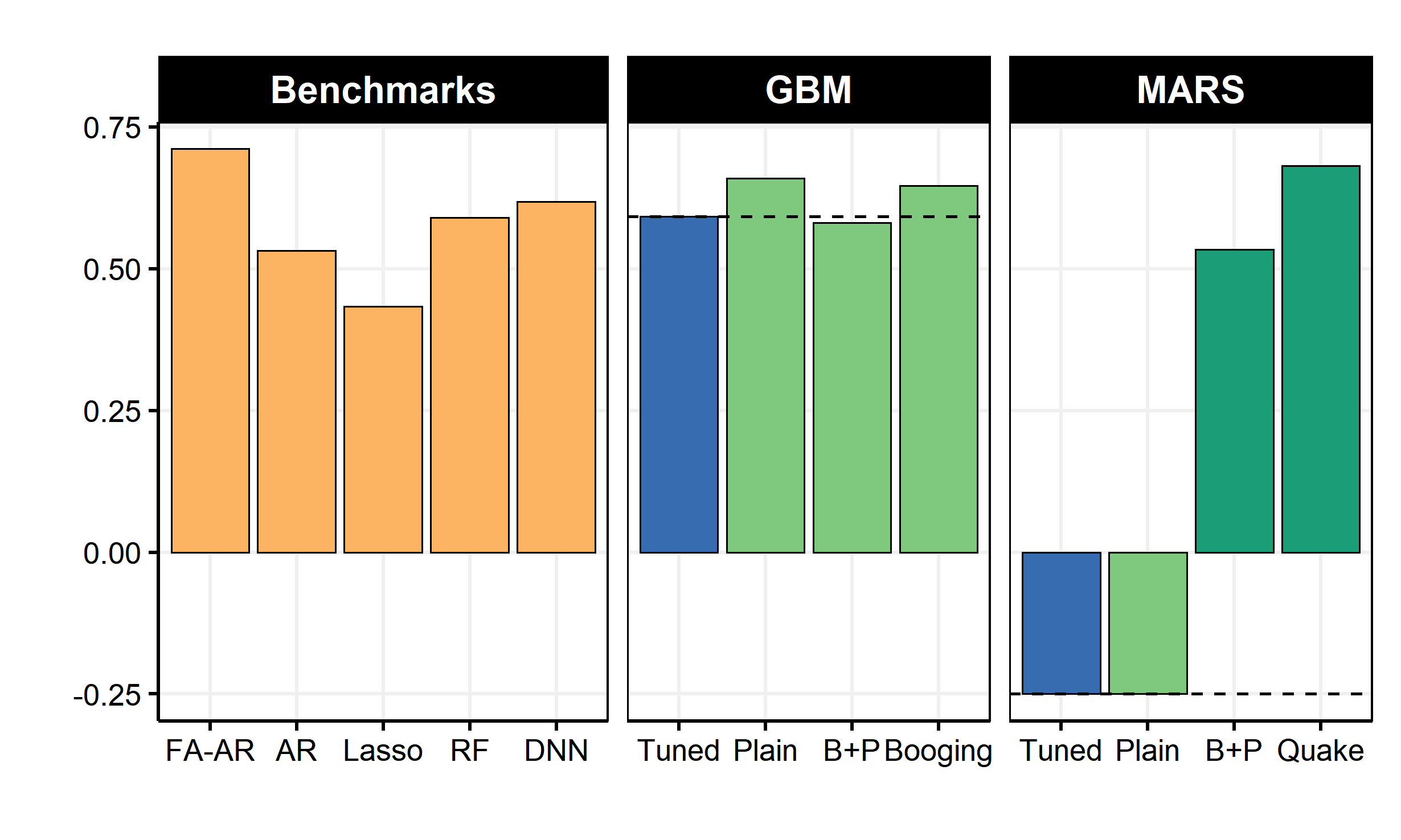}}
\caption{\textit{US Unemployment Rate} $(h=1)$}  
\label{HOUST_EOTB_detail}
     \end{subfigure}
  \begin{subfigure}[b]{0.47\linewidth}
\includegraphics[trim={1cm 1.1cm 0.25cm 1.1cm},clip,scale=.35]{{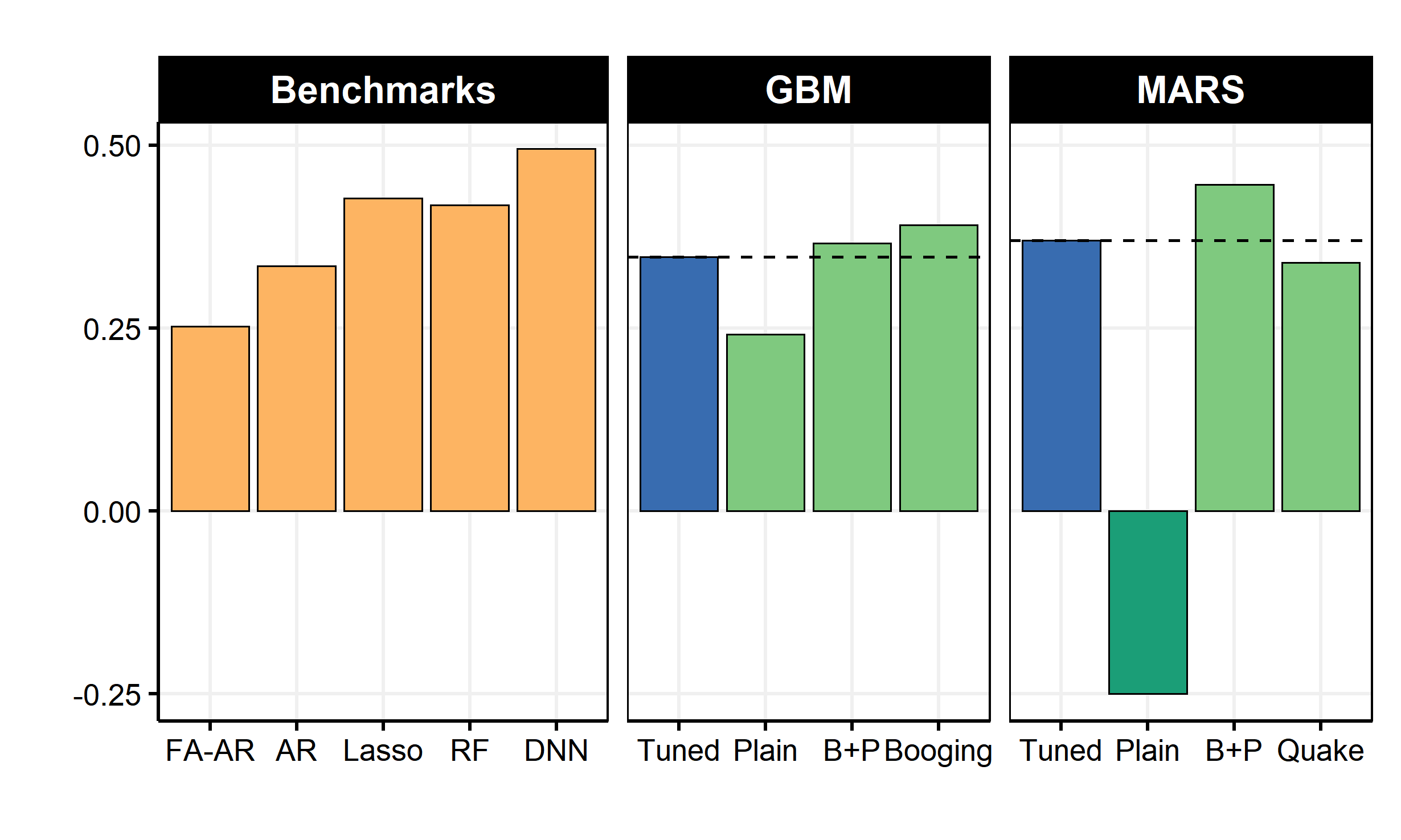}}
\caption{\textit{US Inflation} $(h=1)$}  
\label{IR_EOTB_detail}
      \end{subfigure}
        \begin{subfigure}[t]{0.47\linewidth}
\includegraphics[trim={1cm 1.1cm 0.25cm 1.1cm},clip,scale=.35]{{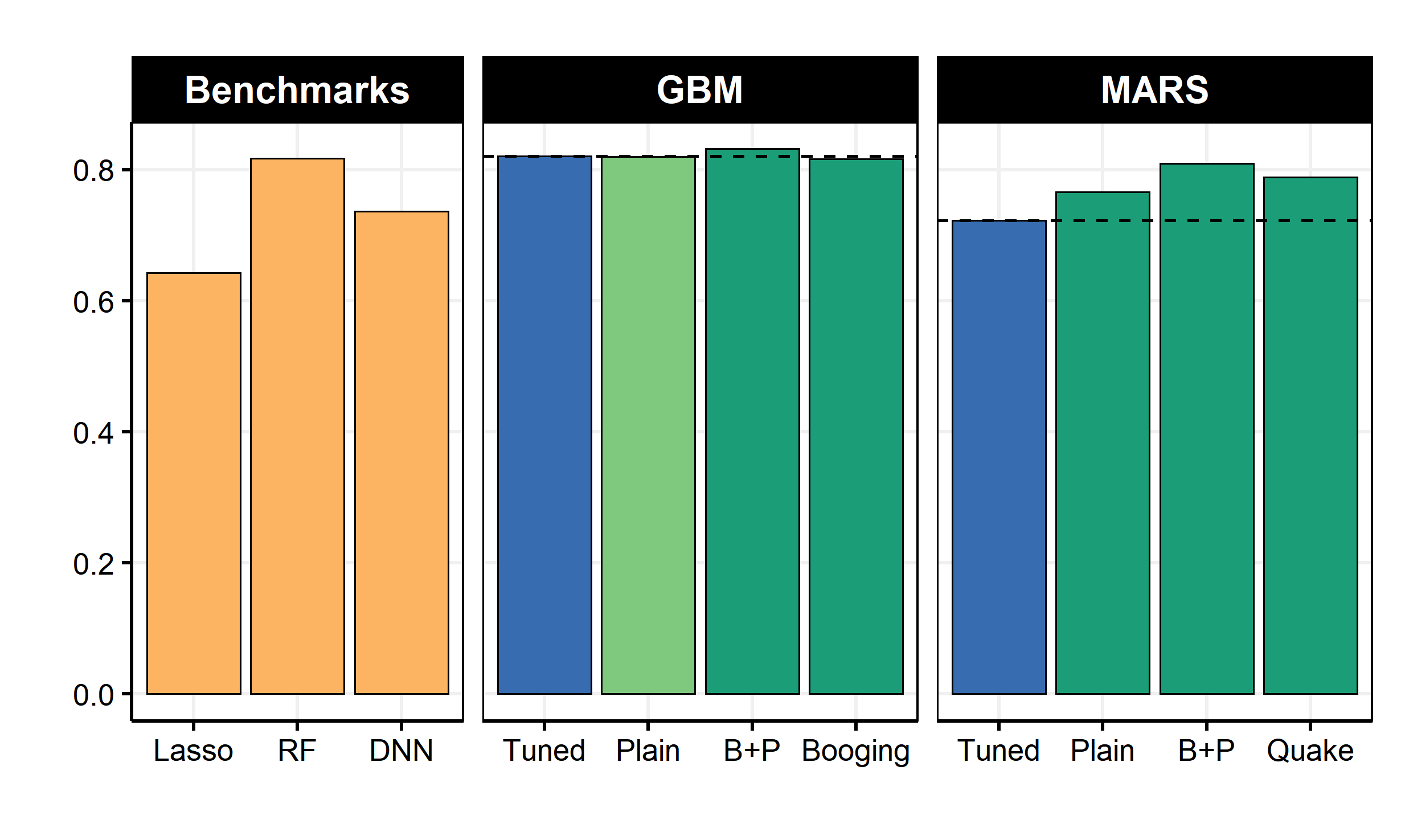}}
\caption{\textit{California Housing}}  
\label{IR_EOTB_detail}
      \end{subfigure}
        \hfill 
        \begin{subfigure}[t]{0.47\linewidth}
\includegraphics[trim={1cm 1.1cm 0.25cm 1.1cm},clip,scale=.35]{{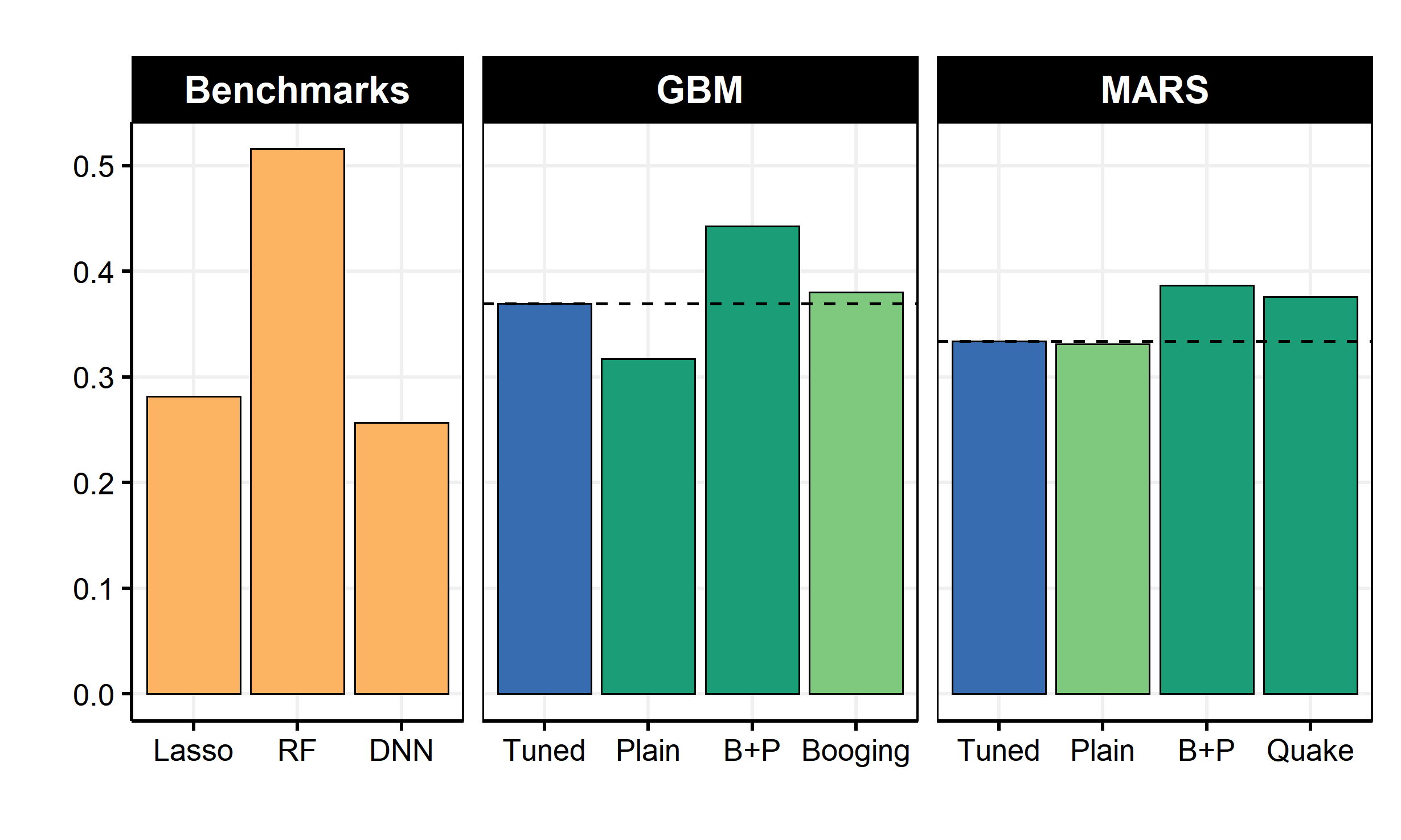}}
\caption{\textit{White Wine}}  
\label{IR_EOTB_detail}
      \end{subfigure}
      \vspace{0.5cm}
  \caption{\footnotesize A Subset of Empirical Prediction Results. Performance metric $R^2_{\text{test}}$. Darker green bars means the performance differential between the tuned version and the three others is statistically significant at the 5\% level using t-tests (and  \cite{dieboldmariano} tests for time series data). Light green means the difference is not significant at the prescribed level. To enhance visibility in certain cases, $R^2_{\text{test}}$'s below -0.25 are constrained to 0.25.}
   \label{results_fig}
\end{figure}

All prediction results are reported in Table \ref{results1} and an example is plotted in Figure \ref{results_fig}. Moreover, to empirically document the $R^2_{\text{test}}$ and $R^2_{\text{train}}$ gap, Table \ref{results1_in} (Appendix) reports $R^2_{\text{train}}$'s. Overall, empirical results confirm the insights developed in section \ref{sec:rf_miracle}. In almost every instance, the overfitting ensembles do at least as well as the tuned version while completely overfitting the training sample, the same way RF would. Sometimes they do better. When they do not, going from  \texttt{B} \& \texttt{P}  to MARSquake and Booging helps. This seldom occurring wedge suggests an important role for data augmentation when features are scarce. 
 
%Thus, they are alternatives to their cross-validated counterparts. 
 
For instance, on the \textit{Abalone} data set, non-tuned MARS is overfitting, which leads to subpar performance. In line with simulation results, the newly proposed overfitting ensembles perform similarly well to using a single base learner and tuning it. Even better, Booging delivers statistically significant \textit{gains} at the 1\% confidence level. As RF, those two ensembles have a very high $R^2_{\text{train}}$ (see Table  \ref{results1_in} in the appendix) and yet, stellar performance is reported on the test set. For \textit{Fish Toxicity}, \textit{Red Wine}, \textit{White Wine}, the plain overfitting versions are significantly worse than the tuned versions, and ensembling them delivers a performance (with respect to tuned counterparts) that is either significantly better or statistically indistinguishable. For \textit{California Housing}, which has more than 20\,000 observations, all ensembles significantly improve over the tuned version for MARS. % and performances are similar within the GBM group. 

For \textit{Crime Florida} -- the very high-dimensional case which is not time series --  the two ensembles of completely overfitting MARS (their $R^2_{\text{train}}$ are respectively 0.97 and 0.98) are doing much better than the tuned version. They both deliver a $R^2_{\text{test}}$ of almost 0.8 in the case of MARSquake. The latter is also the overall second best model (being 1\% less than NN) for this data set. Meanwhile, \texttt{B} \& \texttt{P} Boosting and Booging are doing marginally better than the tuned version.

A now-familiar pattern is also visible for both unemployment and (to some extent) GDP at $h=1$. Booging does as well as the tuned Boosting. Moreover, the former provides the best outcome among all models, with a 11\% $R^2_{\text{test}}$ increase with respect to both economic forecasting workhorses (AR, FA-AR). When it comes to plain and tuned MARS, all models are somewhat worse than the benchmarks with the tuned model itself delivering a terrible $R^2_{\text{test}}$. MARSquake is partially exempted from this failure for GDP, and completely is for unemployment. In the latter case, MARSquake is as good as FA-AR which incredible resilience is vastly documented \citep{SW2002,GCLSS2018}.  For inflation ($h=1$), the best models are clearly \texttt{B} \& \texttt{P} MARS and DNN.  Finally, it is noteworthy that Booging dominates its tuned counterpart for all economic data sets.  Thus, overfitting ensembles work well for economic forecasting where CV can be hazardous.

Lastly, on NN and Deep NN performances. DNN is mostly dominated by RF and other ensembles, with the exception of inflation where it narrowly beats \texttt{B} \& \texttt{P} MARS. Giving the ongoing discussion on the properties of DNN's, it is interesting to check if DNN behaves similarly to RF. The short answer is "no". RF's $R^2_{\text{train}}$ is \textit{almost always} above 0.9, whereas that of DNN fluctuates highly depending on the target. Also, DNN does not display RF's emblematic resilience across data sets.
 
\section{Conclusion}\label{sec:con}
A fundamental problem is to detect at which point a learner stops learning and starts imitating. In ML, the common tool to prevent an algorithm from damaging its hold-out sample performance by overfitting is cross-validation. It is widespread knowledge that performing CV on Random Forests rarely yields dramatic improvements. Concurrently, it is often observed that $R^2_{\text{test}}< R^2_{\text{train}}$ without $R^2_{\text{test}}$ being any less competitive. I argued that proper inner randomization as generated by bagging and perturbing the model, when combined with a greedy fitting procedure, will implicitly prune the learner once it starts fitting noise. By the virtues of recursive model building, the earlier fitting steps are immune to the instability brought upon by subsequent (and potentially harmful) steps.  Once upon a time, the author heard a very senior data scientist and researcher say in a seminar, 'If you put a gun to my head and say "predict", I use Random Forest.' This paper rationalizes this feeling of security by noting that unlike other learners, RF performs its own pruning without the perils of cross-validation. Thus, it seems that, mixed with a proper amount of randomization, \textit{greed is good}.

%The author once heard a very senior data scientist say in a seminar, 'If you put a gun to my head and say "predict", I use Random Forest.' This paper rationalizes the feeling of security by noting that unlike any other ML algorithms, RF performs its own pruning without the perils of cross-validation. 

%\epigraph{}{--- \textup{The Unknown Seminar Participant}}

\clearpage

\setlength\bibsep{5pt}
               
\bibliographystyle{apalike}
 
\setstretch{0.75}
 
%\bibliography{C:/Users/Public/Bibtex_files/ref_pgc_v181204}
%\bibliography{/Users/UQAM/Dropbox/pgc_bib/ref_pgc_v181204}
 
 \bibliography{ref_pgc_v181204}

\clearpage
 
\appendix
%\appendixpage
%\addappheadtotoc
\newcounter{saveeqn}
\setcounter{saveeqn}{\value{section}}
\renewcommand{\theequation}{\mbox{\Alph{saveeqn}.\arabic{equation}}} \setcounter{saveeqn}{1}
\setcounter{equation}{0}
\setstretch{1.25}
 
\pagebreak
 
%%%%%%%%%%%%%%%%%%%%%%%%%%%%%%%%%%%%%%%%%%%%%%%%%%%%%%%%%%%%
 
\appendix
 
\section{Appendix}
\subsection{Additional Graphs and Tables}
 
\begin{figure}[ht!]
\begin{center} %\hspace*{-0.75cm}
\hspace*{-0.4cm}\includegraphics[scale=.45]{{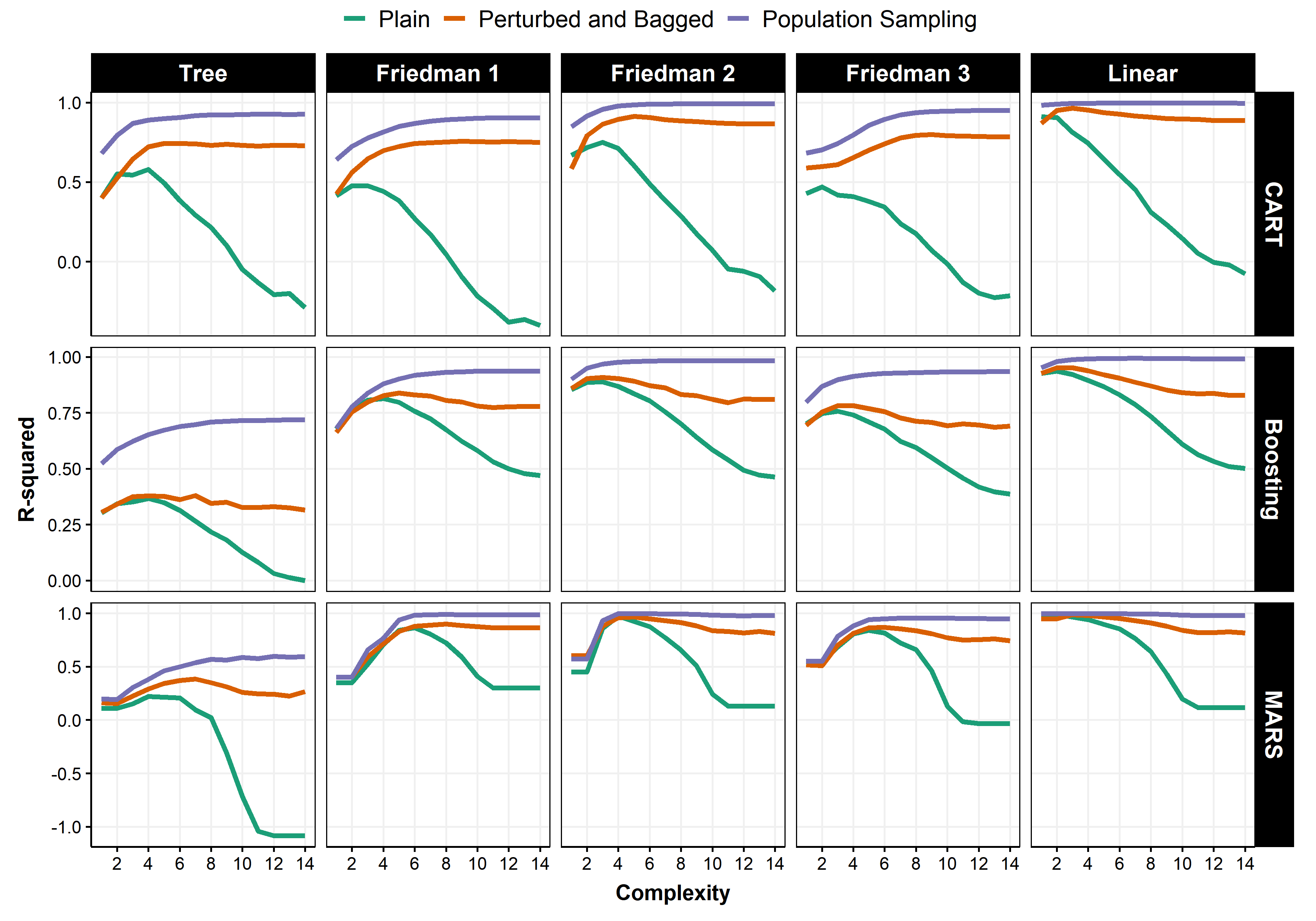}}
\vspace*{-0.4cm}  
\caption{\scriptsize This plots the hold-out sample $R^2$ between the prediction and the true conditional mean. The level of noise is calibrated so the signal-to-noise ratio is 1. Column facets are DGPs and row facets are base learners. The $x$-axis is an index of depth of the greedy model. For CART, it is a decreasing minimal size node $\in 1.4^{\{16,..,2\}}$, for Boosting, an increasing number of steps $\in 1.5^{\{4,..,18\}}$ and for MARS, it is an increasing number of included terms $\in 1.4^{\{2,..,16\}}$.}
\label{simul_nl1}
\end{center}
\end{figure}
 
\begin{figure}[ht!]
\begin{center} %\hspace*{-0.75cm}
\hspace*{-0.3cm}\includegraphics[scale=.44]{{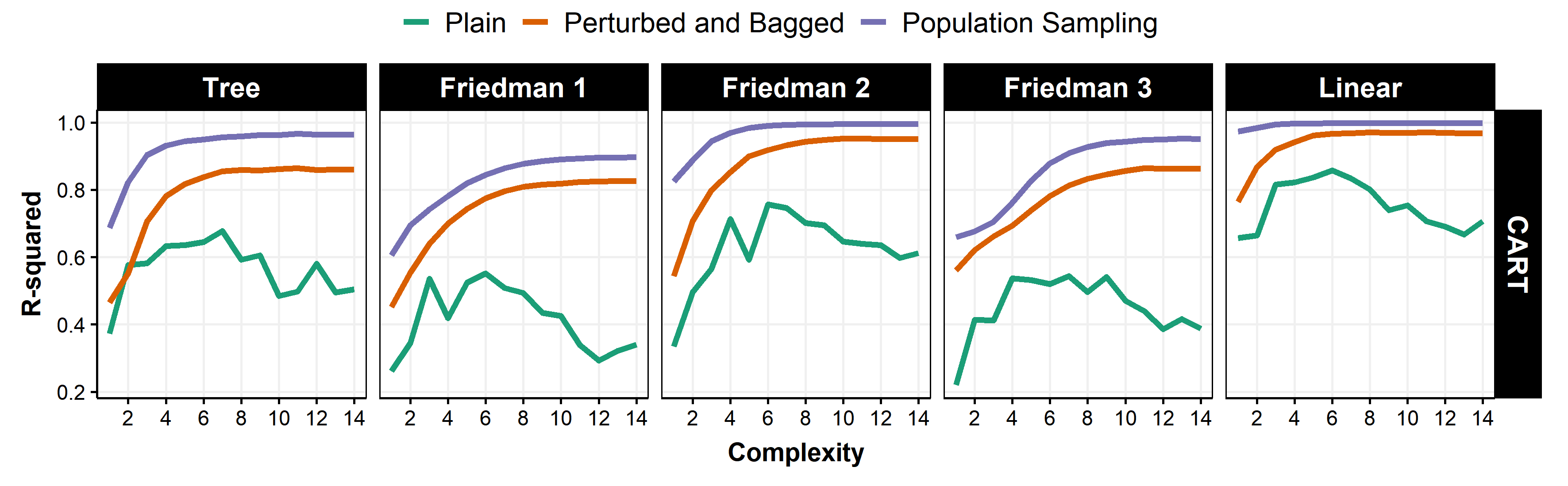}}
\vspace*{-0.4cm}  
\caption{\footnotesize This is Figure \ref{simul_nl2}'s first row with \texttt{mtry}= 0.5.}
\label{simul_nl2_mtry50}
\end{center}
\end{figure}
 
\begin{table}[ht!]
  \caption{20 Data Sets \label{tab_datasets}}
  \centering
  \footnotesize
    \begin{threeparttable}
 
    \begin{tabularx}{0.8\textwidth}{XXXX}
    \toprule \toprule
    \textbf{Abbreviation} & \textbf{Observations} & \textbf{Features} & \textbf{Data Source}\\
    \midrule
Abalone & 4,177 & 7 & \href{http://archive.ics.uci.edu/ml/datasets/Abalone}{archive.ics.uci.edu} \\    
 
Boston Housing & 506 & 13 & \href{http://lib.stat.cmu.edu/datasets/boston}{lib.stat.cmu.edu} \\
     
Auto & 392 & 7 & \href{https://archive.ics.uci.edu/ml/datasets/Auto+MPG}{archive.ics.uci.edu} \\
     
Bike Sharing & 17,379  & 13 & \href{https://archive.ics.uci.edu/ml/datasets/Bike+Sharing+Dataset}{archive.ics.uci.edu}\\
 
White Wine & 4,898 & 10 & \href{https://archive.ics.uci.edu/ml/datasets/Wine+Quality}{archive.ics.uci.edu}  \\
 
Red Wine & 1,599 & 10 & \href{https://archive.ics.uci.edu/ml/datasets/Wine+Quality}{archive.ics.uci.edu}  \\
 
Concrete & 1,030 & 8 & \href{https://archive.ics.uci.edu/ml/datasets/Concrete+Compressive+Strength}{archive.ics.uci.edu}  \\
 
Fish Toxicity & 908 & 6 & \href{https://archive.ics.uci.edu/ml/datasets/QSAR+fish+toxicity}{archive.ics.uci.edu}  \\
 
Forest Fire & 517 & 12 & \href{https://archive.ics.uci.edu/ml/datasets/Forest+Fires}{archive.ics.uci.edu} \\
 
NBA Salary & 483 & 25 & \href{https://www.kaggle.com/aishjun/nba-salaries-prediction-in-20172018-season/data}{kaggle.com} \\
 
CA Housing & 20,428 & 9 & \href{https://www.kaggle.com/camnugent/california-housing-prices}{kaggle.com} \\
 
Crime Florida & 90 & 97 & \href{https://www.census.gov/data/datasets/1990/dec/summary-file-3.html}{census.gov}  \\
 
Friedman 1 $R^2$=.7 & 1,000 & 10 & \href{https://cran.r-project.org/web/packages/tgp/vignettes/tgp.pdf}{cran.r-project.org}   \\
 
Friedman 1 $R^2$=.4 & 1,000 & 10 & \href{https://cran.r-project.org/web/packages/tgp/vignettes/tgp.pdf}{cran.r-project.org}   \\

GDP $h$=1 & 212 & 599 & \href{https://drive.google.com/file/d/1NxDKvr-gyf2hhZS7oJ2h12vFk-N3SZAJ/view?usp=sharing}{Google Drive}  \\
 
GDP $h$=2 & 212 & 563 & \href{https://drive.google.com/file/d/1NxDKvr-gyf2hhZS7oJ2h12vFk-N3SZAJ/view?usp=sharing}{Google Drive}  \\
 
UNRATE $h$=1 & 212 & 619 & \href{https://drive.google.com/file/d/1NxDKvr-gyf2hhZS7oJ2h12vFk-N3SZAJ/view?usp=sharing}{Google Drive} \\
 
UNRATE $h$=2 & 212 & 627 & \href{https://drive.google.com/file/d/1NxDKvr-gyf2hhZS7oJ2h12vFk-N3SZAJ/view?usp=sharing}{Google Drive} \\
 
INF $h$=1 & 212 & 619 & \href{https://drive.google.com/file/d/1NxDKvr-gyf2hhZS7oJ2h12vFk-N3SZAJ/view?usp=sharing}{Google Drive} \\
 
INF $h$=2 & 212 & 611 & \href{https://drive.google.com/file/d/1NxDKvr-gyf2hhZS7oJ2h12vFk-N3SZAJ/view?usp=sharing}{Google Drive} \\
 
\bottomrule \bottomrule
    \end{tabularx}%  
  \begin{tablenotes}[para,flushleft]
  Notes: The number of features includes categorical variables expanded as multiple dummies and will thus be sometimes higher than what reported at data source website. Data source URLs are visibly abbreviated but lead directly to the exact data set or package being used. The number of features varies for each macro data set because a mild screening rule was implemented ex-ante, the latter helping to decrease computing time.
  \end{tablenotes}
    \end{threeparttable}
 
\end{table}%
 
%\pagebreak
 
%\begin{landscape}
%\vspace*{\fill}
%\begin{adjustbox}{center}
 
%\sisetup{group-separator = {,},
%    input-decimal-markers = {.},
%    input-symbols = ()}
 
%\end{adjustbox}
%\vspace*{\fill}
%\end{landscape}

%> load("C:/Users/Philippe GC/Dropbox/Amnesic_NN/Copy_HPCC/boog23nn_newcode_DS28.RData") for NNs
%path='C:/Users/Philippe GC/Dropbox/RF_macro/MSoRF/hpcc_output/boog13_clean/' #_akshay for the others
 
%\pagebreak
\pagebreak

\begin{landscape}
\vspace*{\fill}

%\begin{sidewaystable}[htp]
   % \centering
%\begin{adjustbox}{center}
 
%\sisetup{group-separator = {,},
%    input-decimal-markers = {.},
%    input-symbols = ()}
\begin{table}[htp]
  \begin{threeparttable}
\centering
\footnotesize
\caption{$R^2_{\text{test}}$ for all data sets and models \label{results1}}
\setlength{\tabcolsep}{0.15em}
  \rowcolors{2}{white}{gray!15}
%\sisetup{detect-weight=true,detect-inline-weight=math}
\sisetup{detect-weight, mode=math}
\begin{tabular}{ l *{16}
S[
    table-format            = -2.3,
    input-close-uncertainty = ,
    input-open-uncertainty =  ,
    table-space-text-pre   = ( , % )
    table-space-text-post  = \stars{***},
    table-align-text-post  = false
  ]{c} }
\toprule
\hspace*{0.5cm} &
\multicolumn{7}{c}{Benchmarks} &
\multicolumn{4}{c}{GBM} &
\multicolumn{4}{c}{MARS} \\
\cmidrule(lr){2-8} \cmidrule(lr){9-12} \cmidrule(lr){13-16}
&
{FA-AR} & %\makebox[3em]
{AR} &  %\makebox[3em]
{LASSO} &
{RF} &
{Tree} &
{NN} &
{DNN} &
{Tuned} &
{Plain} &
{\texttt{B} \& \texttt{P}} &
{Booging} &
{Tuned} &
{Plain} &
{\texttt{B} \& \texttt{P}} &
{Quake}\\
\midrule
Abalone&  & & 0.52& 0.56&0.45& 0.54 & 0.53& 0.50& 0.48& 0.53*& 0.54**& 0.57& 0.35*& 0.31*&\bfseries 0.58***\tabularnewline
Boston Housing&  & & 0.67& 0.88&0.79& 0.86 & 0.85& 0.89& 0.88& 0.90& 0.85*& 0.83& 0.87&\bfseries 0.92& 0.91\tabularnewline
Auto&  & & 0.66&\bfseries 0.71&0.61&  0.13& 0.64& 0.64& 0.59$^{**}$& 0.65& 0.64*& 0.71&-0.54*& 0.53& 0.63\tabularnewline
Bike Sharing&  & & 0.38& 0.91&0.73& 0.88 & 0.94&\bfseries 0.95& 0.93***& 0.91***& 0.91***& 0.71& 0.89***& 0.87***& 0.90***\tabularnewline
White Wine&  & & 0.28&\bfseries 0.52&0.28& 0.37 &0.26 & 0.37& 0.32*& 0.44***& 0.38& 0.33& 0.33***& 0.39**& 0.38***\tabularnewline
Red Wine&  & & 0.34&\bfseries 0.47&0.35& 0.33 &0.37 & 0.37& 0.23**& 0.37& 0.38& 0.38& 0.29*& 0.33& 0.35\tabularnewline
Concrete&  & & 0.59& 0.90&0.71& 0.89 & 0.88 &\bfseries 0.92& 0.92& 0.90*& 0.90***& 0.83& 0.87& 0.30***& 0.89\tabularnewline
Fish Toxicity&  & & 0.56&\bfseries 0.65&0.57& 0.60 & 0.63& 0.63& 0.54***& 0.61& 0.62& 0.56&-0.25***& 0.54*& 0.61\tabularnewline
Forest Fire&  & & 0.00&-0.11&0.00& -0.02 & 0.01&-0.03&-0.68***&-0.32***&-0.08&\bfseries 0.01&-1.55*&-0.68&-0.36\tabularnewline
NBA Salary&  & & 0.52&\bfseries 0.60&0.34&  0.22& 0.21& 0.50& 0.29***& 0.49& 0.50& 0.36& 0.11*& 0.59*& 0.53\tabularnewline
CA Housing&  & & 0.64& 0.82&0.59& 0.75 & 0.74& 0.82& 0.82&\bfseries  0.83***& 0.82**& 0.72& 0.77***& 0.81***& 0.79***\tabularnewline
Crime Florida&   &  &0.66& 0.79&0.60& \bfseries 0.82 & 0.75& 0.75& 0.77& 0.81*& 0.79& 0.70& 0.44*& \bfseries 0.81& 0.80\tabularnewline
F1 $R^2=0.7$&  & & 0.53& 0.62&0.50& 0.43 &0.51 & 0.65& 0.54***& 0.60***& 0.67**& 0.68& 0.55& 0.62&\bfseries 0.69***\tabularnewline
F1 $R^2=0.4$ &  & & 0.32& 0.40&0.36& 0.19 & 0.28& 0.40& 0.16***& 0.34*& 0.41&\bfseries 0.41& 0.14*& 0.35& 0.40*\tabularnewline
GDP $h$=1& 0.27&0.27& 0.24& 0.35&0.18& 0.06 &0.26 & 0.36& 0.17& 0.37&\bfseries 0.38& 0.00&-9.08***&-0.45**&-0.12**\tabularnewline
GDP $h$=2&-0.03&0.17&-0.01& 0.16&0.00& -0.06 & -0.52& 0.15&-0.56**&\bfseries 0.20\bfseries& 0.18&-0.40&-4.37**&-0.41*&-0.37***\tabularnewline
UNRATE $h$=1&\bfseries 0.71&0.53& 0.43& 0.59&0.22&-0.69 &0.62 & 0.59& 0.66& 0.58& 0.65&-0.65&-0.72***& 0.53& 0.68\tabularnewline
UNRATE $h$=2&\bfseries 0.52&0.29& 0.26& 0.37&0.16& 0.14 &0.41 & 0.43& 0.35& 0.42& 0.48& 0.16&-0.80**&-0.28& 0.26\tabularnewline
INF $h$=1& 0.25&0.33& 0.43& 0.42&0.25& 0.41 &\bfseries 0.49 & 0.35& 0.24& 0.37& 0.39& 0.37&-0.57**& 0.45 & 0.34\tabularnewline
INF $h$=2& 0.05&0.22& 0.09& 0.28& 0.45& 0.19 &\bfseries 0.51 & 0.15&-0.26***& 0.16& 0.27*& 0.39&-2.50**& 0.24& 0.42\tabularnewline\bottomrule
\end{tabular}
\begin{tablenotes}[para,flushleft]
  Notes: This table reports $R^2_{\text{test}}$ for 20 data sets and different models, either standard or introduced in the text. For macroeconomic targets (the last 6 data sets), the set of benchmark models additionally includes an autoregressive model of order 2 (AR) and a factor-augmented regression with 2 lags (FA-AR). Numbers in bold identify the best predictive performance of the row. For GBM and MARS, t-test (and  \cite{dieboldmariano} tests for time series data) are performed to evaluate whether the difference in predictive performance between the tuned version and the remaining three models of each block is statistically significant. '*', '**' and '***' respectively refer to p-values below 5\%, 1\% and 0.1\%. F1 means "Friedman 1" DGP of \cite{MARS}.
  \end{tablenotes}
  \end{threeparttable}
 
\end{table}
%\end{adjustbox}

%\end{sidewaystable}

\vspace*{\fill}
\end{landscape}

%> load("C:/Users/Philippe GC/Dropbox/Amnesic_NN/Copy_HPCC/boog23nn_newcode_DS28.RData") for NNs
%path='C:/Users/Philippe GC/Dropbox/RF_macro/MSoRF/hpcc_output/boog13_clean/' #_akshay for the others
 
\pagebreak

%\begin{comment}
        \begin{landscape}
        \vspace*{\fill}
       % \begin{adjustbox}{center}

\begin{table}[htp]
%\begin{sidewaystable}[htp]

  \begin{threeparttable}
\centering
\footnotesize
 
\caption{$R^2_{\text{train}}$ for all data sets and models \label{results1_in}}
\setlength{\tabcolsep}{0.5em}
  \rowcolors{2}{white}{gray!15}
\begin{tabular}{ l *{16}{c} }
\toprule
\rowcolor{white}
\hspace*{0.5cm} &
\multicolumn{7}{c}{Benchmarks} &
\multicolumn{4}{c}{GBM} &
\multicolumn{4}{c}{MARS} \\
\rowcolor[gray]{0}
\cmidrule(lr){2-8} \cmidrule(lr){9-12} \cmidrule(lr){13-16}
&
{FA-AR} & %\makebox[3em]
{AR} &  %\makebox[3em]
{LASSO} &
{RF} &
{Tree} &
{NN} &
{DNN} &
{Tuned} &
{Plain} &
{\texttt{B} \& \texttt{P}} &
{Booging} &
{Tuned} &
{Plain} &
{\texttt{B} \& \texttt{P}} &
{Quake}\\
\midrule
Abalone &   & &0.50&0.92&0.50 & 0.60 & 0.59 &0.53&0.85&0.86&0.91&0.57&0.65&0.78&0.61\tabularnewline
Boston Housing &   & &0.72&0.98&0.87&  0.90 &  0.89 &1.00&1.00&0.99&0.99&0.90&0.97&0.97&0.98\tabularnewline
Auto &   &  &0.68&0.96&0.77& 0.13  & 0.81 &0.86&1.00&0.98&0.98&0.77&0.98&0.93&0.96\tabularnewline
Bike Sharing &   &  &0.38&0.98& 0.89  & 0.95 &0.96&0.95&0.94&0.95&0.71&0.89&0.88&0.90\tabularnewline
White Wine &   &   &0.26&0.92&0.27& 0.47 & 0.75 &0.44&0.82&0.85&0.88&0.37&0.46&0.52&0.51\tabularnewline
Red Wine &   &   &0.29&0.91&0.41& 0.40 &0.42 &0.41&0.96&0.94&0.95&0.44&0.56&0.69&0.67\tabularnewline
Concrete &   &   &0.61&0.98&0.75& 0.91 & 0.93 &0.98&0.99&0.98&0.99&0.88&0.98&0.74&0.95\tabularnewline
Fish Toxicity &   &   &0.54&0.93&0.60& 0.64 & 0.61 &0.92&0.97&0.95&0.97&0.63&0.96&0.82&0.88\tabularnewline
Forest Fire &   &   &0.00&0.81&0.00&  0.00&0.07 &0.40&0.97&0.88&0.91&0.04&0.62&0.73&0.76\tabularnewline
NBA Salary &   &   &0.47&0.93&0.72& 0.65 &0.71 &0.99&1.00&0.97&0.97&0.64&0.92&0.84&0.93\tabularnewline
CA Housing &   &   &0.63&0.97&0.61&  0.78&0.85 &0.86&0.89&0.91&0.90&0.72&0.80&0.83&0.81\tabularnewline
Crime Florida &   &   &0.65&0.96&0.84& 0.88 &0.94 &1.00&1.00&0.98&0.98&0.75&1.00&0.97&0.98\tabularnewline
F1 $R^2=0.7$ &   &   &0.45&0.93&0.45& 0.62 & 0.71 &0.95&1.00&0.97&0.97&0.65&0.81&0.84&0.86\tabularnewline
F1 $R^2=0.4$  &   &   &0.23&0.89&0.30& 0.34 & 0.35&0.48&1.00&0.94&0.94&0.38&0.64&0.75&0.76\tabularnewline
GDP $h$=1 &0.41&0.11&0.23&0.91&0.51& 0.26 & 0.44 &0.81&1.00&0.96&0.96&0.47&1.00&0.94&0.94\tabularnewline
GDP $h$=2 &0.26&0.06&0.07&0.89&0.00& 0.26 & 0.55 &0.76&1.00&0.95&0.95&0.29&1.00&0.94&0.95\tabularnewline
UNRATE $h$=1 &0.57&0.40&0.48&0.93&0.81& -0.07 & 0.82 &0.83&1.00&0.97&0.97&0.76&0.99&0.97&0.96\tabularnewline
UNRATE $h$=2 &0.41&0.13&0.35&0.92&0.38& 0.42 &0.25 &0.99&1.00&0.96&0.96&0.75&1.00&0.96&0.96\tabularnewline
INF $h$=1 &0.76&0.73&0.90&0.97&0.81& 0.64 & 0.94 &1.00&1.00&0.99&0.99&0.73&1.00&0.99&0.99\tabularnewline
INF $h$=2&0.69&0.63&0.72 &0.96&0.72& 0.67 & 0.92 &1.00&1.00&0.99&0.98&0.81&1.00&0.99&0.98\tabularnewline
\bottomrule
\end{tabular}
\begin{tablenotes}[para,flushleft]
  Notes: This table reports $R^2_{\text{train}}$ for 20 data sets and different models, either standard or introduced in the text. For macroeconomic targets (the last 6 data sets), the set of benchmark models additionally includes an autoregressive model of order 2 (AR) and a factor-augmented regression with 2 lags (FA-AR). F1 means "Friedman 1" DGP of \cite{MARS}. %Numbers in bold identifies the best predictive performance of the row. For GBM and MARS, t-test (and  \cite{dieboldmariano} tests for time series data) are performed to evaluate whether the difference in predictive performance between the tuned version and the remaining three models of each block is statistically significant. '*', '**' and '***' respectively refer to p-values below 5\%, 1\% and 0.1\%.
  \end{tablenotes}
  \end{threeparttable}
 
\end{table}
%\end{sidewaystable}

%\end{adjustbox}
\vspace*{\fill}
 
\end{landscape}
%\end{comment}
 
%\subsection{MARS and Boosting Overview}
 
 \clearpage
 
\subsection{Implementation Details for \textit{Booging} and \textit{MARSquake}}\label{sec:details}
 
Booging and MARSquake are the \texttt{B} \& \texttt{P} +\texttt{DA} versions of Boosted Trees and MARS, respectively. The data-augmentation option will likely be redundant in high-dimensional situations where the available regressors already have a factor structure (like macroeconomic data). 
\vskip 0.2cm
 
{\sc \noindent \textbf{About \texttt{B}}.} For both algorithms, \texttt{B} is made operational by subsampling. As usual, reasonable candidates for the sampling rate are $\sfrac{2}{3}$ and $\sfrac{3}{4}$. All ensembles use $B=100$ subsamples.
 
\vskip 0.2cm
 
{\sc \noindent \textbf{About \texttt{P}}.} The primary source of perturbation in Booging is straightforward. Using subsamples to construct trees at each step is already integrated within Stochastic Gradient Boosting. By construction, it perturbs the Boosting fitting path and achieve a similar goal as that of the original \texttt{mtry} in RF. Note that, for fairness, this standard feature is also activated for any reported results on "plain" Boosting. %Another layer of randomization can be activated, especially when \texttt{DA} is being used. 
 
The implementation of \texttt{P} in MARSquake is more akin to that of RF. At each step of the forward pass, MASS evaluate all variables as potential candidates to enter a hinge function, and select the one which (greedily) maximize fit at this step. In the spirit of RF's \texttt{mtry}, \texttt{P} is applied by stochastically restricting the set of available features at each step. I set the fraction of randomly considered $X$'s to $\sfrac{1}{2}.$ 
 
To further enhance perturbation in both algorithms, we can randomly drop a fraction of features from base learners' respective information sets. Since \texttt{DA} creates replicas of the data and keep some of its correlation structure, features are unlikely to be entirely dropped from a boosting run, provided the dropping rate is not too high. I suggest 20\%. This can is analogous to \texttt{mtry}-like randomly select features, but for a whole tree (in RF) rather than at each split.

\vskip 0.2cm
 
{\sc \noindent \textbf{About \texttt{DA}}.} Perturbation work better if there is a lot to perturb. In many data sets, $X$ is rich in observations but contains few regressors. To assure \texttt{P} meets its full randomization potential, a cheap data augmentation procedure can be carried. \texttt{DA} is simply adding fake regressors that are correlated with the original $X$ and maintain in part their cross-correlation structure.  Say $X$ contains $K$ regressors. I take the $N \times K$   matrix $X$ and create two duplicates $\tilde{X}= X + \mathcal{E}$ where $\mathcal{E}$ is a matrix of Gaussian noise. SD is set to $\sfrac{1}{3}$ that of the variable. For $X_k$'s that are either categorical or ordinal, I create the corresponding $\tilde{X}_k$ by taking $X_k$ and shuffling 20\% of its observations.
 
\vskip 0.2cm
 
{\sc \noindent \textbf{Last Word on MARS}.} It is known that standard MARS has a forward and a backward pass. The latter's role is to prevent overfitting by (traditional) pruning. Obviously, there is no backward pass in MARSquake. Certain implementations of MARS (like \textit{earth}, \cite{earthpackage}) may contain foolproof features rendering the forward pass recalcitrant to blatantly overfit in certain situations (usually when regressor are not numerous). To partially circumvent this rare occurrence, one can run MARS again on residuals obtained from a first MARS run which failed to attain a high enough $R^2_{\text{train}}$. % as the whole point is to build an ensemble that overfits the traning set, but is self-pruning on the test set. 
 
\section{Simulation Details}\label{sec:simdet}
 
\noindent \textbf{Tree:} The tree DGP is constructed as follows.  Normal noise is generated and a CART tree is fitted to it with 10 normal and independant regressor. The minimal node size to consider a split is 100, which is one fourth of the training sample. This typically generates trees of around 8 nodes. The "fake" conditional mean estimated from this procedure itself used to generate data,  on top of which is added two different level of normal noise as described in Figures \ref{simul_nl1} and Figures \ref{simul_nl2}.  Finally, each model fitted on this DGP is given all the original 10 variables, whether they were used or not by the conditional mean function. 
 
\noindent \textbf{Friedman 1:} Inputs are 10 independent variables uniformly distributed on the interval \([0,1]\), only 5 out of these 10 enter the DGP so that $$y_i = 10 \sin(\pi x_{1,i} x_{2,i}) + 20 (x_{3,i} - 0.5)^2 + 10 x_{4,i}+ 5 x_{5,i} + \epsilon_i$$ with $\epsilon_i$ being normal noise.  
 
\noindent For \textbf{Friedman 2} and \textbf{Friedman 3},  regressors are
$x_{1,i} \in [0,100], x_{2,i}\in [40 \pi,560 \pi],x_{3,i}\in[0,1],x_{4,i}\in [1,11]$
and the targets are
$$\text{\textbf{F2}:}  \quad y_i = (x_{1,i}^2 + (x_{2,i} x_{3,i}- (1/(x_{2,i} x_{4,i})))^2)^{0.5} + \epsilon_i $$ 
$$ \text{\textbf{F3}:} \quad  y_i = \text{atan}((x_{2,i} x_{3,i}- (1/(x_{2,i} x_{4,i})))/x_{1,i}) + \epsilon_i $$
with $\epsilon_i$ being normal noise.
 
\noindent \textbf{Linear:} The linear DGP is the sum of the first variables in F1, with normal noise.  The model is fed 10 variables, with 5 being actually useful.
 
\subsection{Additional NN details}\label{sec:nndetails}
 
For both neural networks, the batch size is 32 and the optimizer is
Adam (with Keras default values). Continuous $X$'s are normalized so that all values are within the 0-1 range.  %The final prediction is the average of an ensemble of 5 different estimations so to average out the effect of weights initialization.
 
More precisely, \textbf{NN} in Table \ref{results1} is a standard feed-forward fully-connected network with an architecture in the vein of \cite{kellyml}. There are two hidden layers, the first with 32 neurons and the second with 16 neurons. The
number of epochs is fixed at 100. The activation function is \textit{ReLu} and that
of the output layer is linear.  The learning rate $%
\in \{0.001,0.01\}$ and the LASSO $\lambda$ parameter $\in \{0.001,0.0001\}$ are
chosen by 5-fold cross-validation. A batch normalization layer follows each \textit{ReLu} layers. Early stopping is applied by stopping training whenever 20 epochs pass without any improvement of the
cross-validation MSE. 
 
More precisely, \textbf{DNN} in Table \ref{results1} is a standard feed-forward fully-connected network with an architecture closely
following that of \cite{olson2018modern} for small data sets. There are 10 hidden layers, each featuring 100 neurons. The
number of epochs is fixed at 200. The activation function is \textit{eLu} and that
of the output layer is linear. The learning rate $%
\in \{0.001,0.01,0.1\}$ and the LASSO $\lambda$ parameter $\in \{0.001,0.00001\}$ are
chosen by 5-fold cross-validation. No early stopping is applied.

\end{document}